\documentclass[10pt,twocolumn,letterpaper]{article}

\usepackage{cvpr}              

\usepackage[dvipsnames]{xcolor}

\definecolor{gary}{RGB}{0,120,120}

\usepackage{float}
\usepackage{booktabs}
\usepackage{multirow}
\usepackage{xcolor,colortbl}

\definecolor{cvprblue}{rgb}{0.21,0.49,0.74}
\usepackage[pagebackref,breaklinks,colorlinks,citecolor=cvprblue]{hyperref}

\def\paperID{4748} 
\def\confName{CVPR}
\def\confYear{2024}

\newcommand\blfootnote[1]{%
  \begingroup
  \renewcommand\thefootnote{}\footnote{#1}%
  \addtocounter{footnote}{-1}%
  \endgroup
}

\title{DNGaussian: Optimizing Sparse-View 3D Gaussian Radiance Fields \\ with Global-Local Depth Normalization}

\author{
    Jiahe Li$^1$, Jiawei Zhang$^1$, Xiao Bai$^1$\thanks{Corresponding author: Xiao Bai (baixiao@buaa.edu.cn).}, Jin Zheng$^1$, Xin Ning$^2$, Jun Zhou$^3$, Lin Gu$^{4,5}$ \\
    $^1$School of Computer Science and Engineering, State Key Laboratory of \\ Complex \& Critical Software Environment,\, Jiangxi Research Institute,\, Beihang University\\
    $^2$Institute of Semiconductors, Chinese Academy of Sciences\\
    $^3$School of Information and Communication Technology, Griffith University\\
    $^4$RIKEN AIP \quad$^5$The University of Tokyo
}

\begin{document}
 
\twocolumn[{
\renewcommand\twocolumn[1][]{#1}
\maketitle

\begin{center}
    \vspace{-6mm}
    \setlength{\abovecaptionskip}{4pt}
    \centering
    \includegraphics[width=1.0\linewidth]{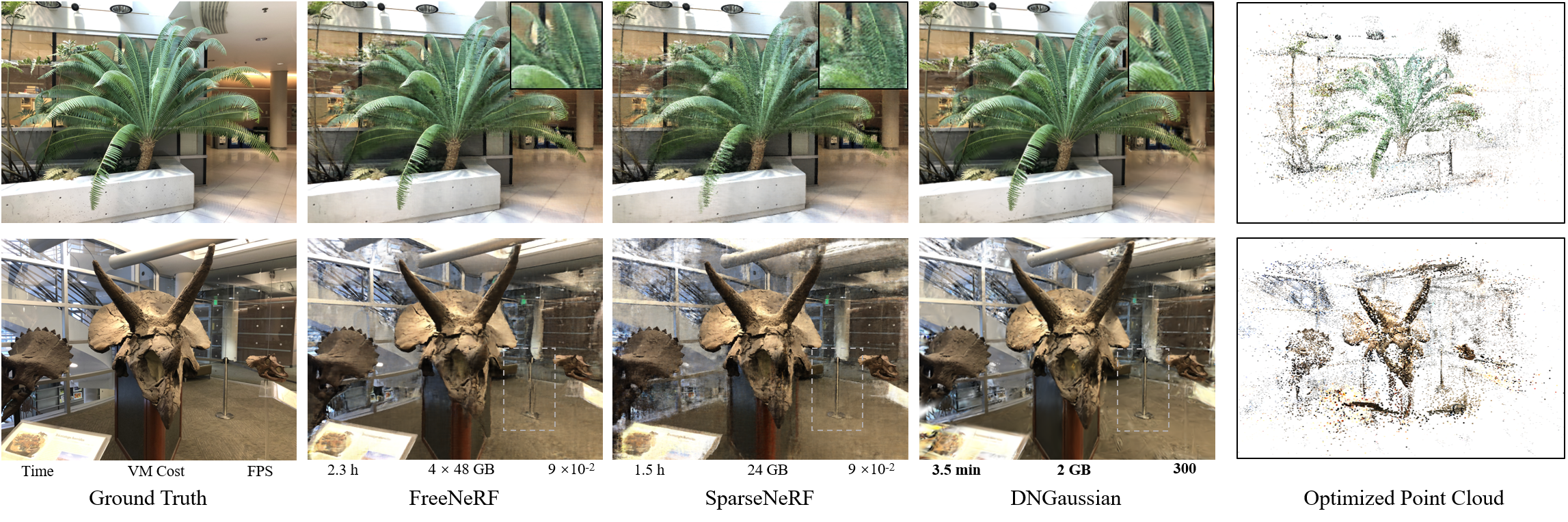}
    \captionof{figure}{Comparison of the state-of-the-arts FreeNeRF \cite{yang2023freenerf} and SparseNeRF \cite{wang2023sparsenerf} with our DNGaussian utilizing three views for training. DNGaussian stands out by delivering comparably high-quality synthesized views and superior details with a remarkable 25× reduction in time and significantly lower memory overhead during training, while attaining the fastest and the only real-time rendering speed of 300 FPS. The point cloud of Gaussians illustrates the detailed and explainable spatial representation learned through our method.} 
    \label{fig:demo1} 
\end{center}
}
]

\blfootnote{$^*$Corresponding author: Xiao Bai (baixiao@buaa.edu.cn).}

\begin{abstract}
Radiance fields have demonstrated impressive performance in synthesizing novel views from sparse input views, yet prevailing methods suffer from high training costs and slow inference speed. This paper introduces DNGaussian, a depth-regularized framework based on 3D Gaussian radiance fields, offering real-time and high-quality few-shot novel view synthesis at low costs. Our motivation stems from the highly efficient representation and surprising quality of the recent 3D Gaussian Splatting, despite it will encounter a geometry degradation when input views decrease. In the Gaussian radiance fields, we find this degradation in scene geometry primarily lined to the positioning of Gaussian primitives and can be mitigated by depth constraint. Consequently, we propose a Hard and Soft Depth Regularization to restore accurate scene geometry under coarse monocular depth supervision while maintaining a fine-grained color appearance. To further refine detailed geometry reshaping, we introduce Global-Local Depth Normalization, enhancing the focus on small local depth changes. Extensive experiments on LLFF, DTU, and Blender datasets demonstrate that DNGaussian outperforms state-of-the-art methods, achieving comparable or better results with significantly reduced memory cost, a $25\times$ reduction in training time, and over $3000\times$ faster rendering speed. 
Code is available at: \url{https://github.com/Fictionarry/DNGaussian} .
\end{abstract}

\vspace{-1mm}
\section{Introduction}

Novel view synthesis with sparse inputs poses a challenge for radiance fields. Recent advances in neural radiance fields (NeRF) have excelled in reconstructing photorealistic appearance and accurate geometry from just a handful of input views \cite{niemeyer2022regnerf, yang2023freenerf, wang2023sparsenerf, seo2023flipnerf, wynn2023diffusionerf, song2023darf, chen2021mvsnerf, yu2021pixelnerf, jain2021putting}. However, most sparse-view NeRFs are implemented with low processing speed and substantial memory consumption, resulting in high time and computational costs that restrict their practical applications. While some methods \cite{song2023darf, sun2023vgos, wynn2023diffusionerf} achieve faster inference speed with grid-based backbones \cite{muller2022instant, sun2022direct, fridovich2023k}, they often suffer from trade-offs, leading to either high training costs or compromised rendering quality. 

\begin{figure}
    \setlength{\abovecaptionskip}{2pt}

    \centering
    \includegraphics[width=\linewidth]{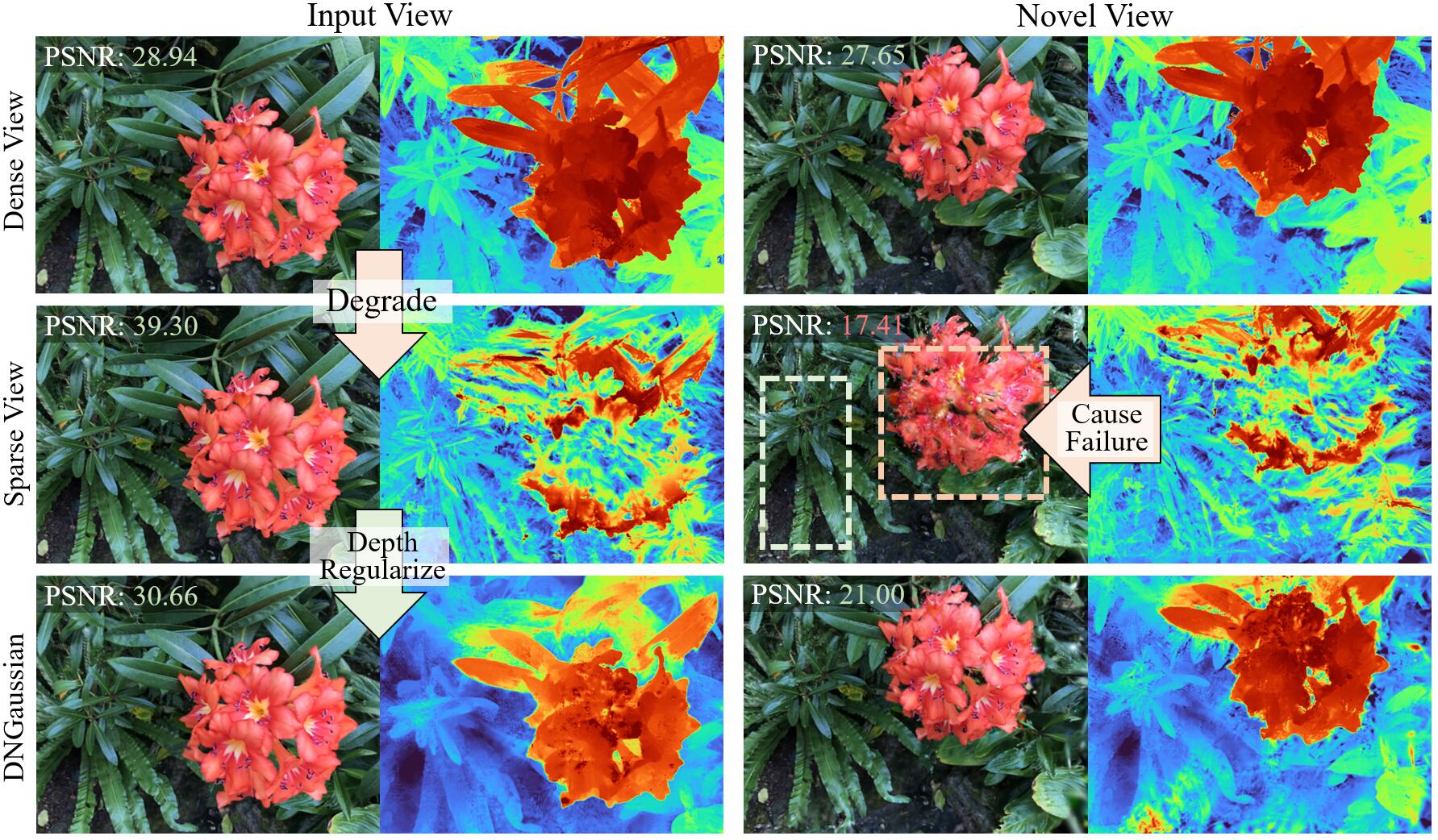}
    \caption{3D Gaussian Splatting \cite{kerbl20233d} exhibits its potential to reconstruct some fine details (green box) from sparse input views. Nevertheless, the reduced input views would significantly degrade geometry and cause failed reconstruction (orange box). After applying depth regularization, DNGaussian successfully recovers accurate geometry and synthesizes high-quality novel views. } 
    \vspace{-16pt}
    \label{fig:illus}
\end{figure}

Recently, 3D Gaussian Splatting \cite{kerbl20233d} has introduced an unstructured 3D Gaussian radiance field, employing a set of 3D Gaussian primitives to achieve remarkable success in rapid, high-quality, and low-cost novel view synthesis, when learned from color dense input views.
Even with only sparse inputs, it can still partially retain the surprising ability to reconstruct some clear and detailed local features. Nevertheless, the decrease in view constraints makes a significant portion of scene geometry be incorrectly learned, resulting in failures in novel view synthesis, as illustrated in Figure~\ref{fig:illus}. Inspired by the success of earlier depth-regularized sparse-view NeRFs~\cite{wang2023sparsenerf, song2023darf}, this paper explores distilling depth information from pre-trained monocular depth estimators to rectify the Gaussian fields of the ill-learned geometry, and introduce the Depth Normalization Regularized Sparse-view 3D Gaussian Radiance Fields (\textbf{DNGaussian}) to pursue higher quality and efficiency for few-shot novel view synthesis.

Despite sharing a similar form of depth rendering, the depth regularization for 3D Gaussian radiance fields differs significantly from that employed by NeRF. Firstly, existing depth regularization strategies for NeRFs commonly employ depth to regularize the entire model, which creates a potential geometry conflict in the Gaussian fields that adversely affects quality. Specifically, this practice forces the shape of Gaussians to fit the smooth monocular depth rather than the complex 
color appearance and thus results in loss of details and blurred appearance. Considering that the basis of scene geometry lies in the position of the Gaussian primitives rather than their shape, we freeze the shape parameters and propose a \textit{Hard and Soft Depth Regularization} to enable spatial reshaping by encouraging movement among the primitives. During regularization, we propose rendering two types of depth to independently adjust the center and opacity of Gaussians without changing their shape, therefore striking a balance between the fitting of complex color appearance and smooth coarse depth.

Moreover, Gaussian radiance fields are more sensitive to small depth errors when compared to NeRF, which may result in a noisy distribution of primitives and failures in regions with complex textures. Existing scale-invariant depth losses often opt to align depth maps to a fixed scale, which leads to the overlook of small losses. To address this issue, we introduce the \textit{Global-Local Depth Normalization} into the depth loss function, thus encouraging the learning of small local depth changes in a scale-invariant way. With the local and global scale normalization, our method guides the loss function to refocus on small local errors while maintaining knowledge on the absolute scale, to enhance the detailed geometry reshaping process for depth regularization.

Integrating the two proposed techniques, DNGaussian synthesizes views with competitive quality and superior details compared to state-of-the-art methods in multiple sparse-view settings on LLFF, Blender, and DTU datasets. This advantage is further enriched by substantially lower memory costs, $25\times$ reduction of training time, and over $3000\times$ faster rendering speed. The experiments also demonstrate our method's universal ability to fit complex scenes, wide-ranging views, and multiple materials.

Our main contributions are the following:

\vspace{-4pt}
\begin{itemize}
    \item A Hard and Soft Depth Regularization to constrain the geometry of 3D Gaussian radiance fields by encouraging the movement of Gaussians, which enables the coarse-depth regularized space reshaping without compromising fine-grained color performance. 
    \item A Global-Local Depth Normalization that normalizes depth patches on local scales to achieve a refocus on small local depth changes, thereby improving the reconstruction of detail appearance for 3D Gaussian radiance fields. 
    \item A DNGaussian framework for fast and high-quality few-shot novel view synthesis, which combines the above two techniques and achieves competitive quality across multiple benchmarks compared to the state-of-the-art methods, excelling in capturing details with significantly lower training costs and real-time rendering.
\end{itemize}

\vspace{-4pt}\noindent 
To the best of our knowledge, we are the first attempt to analyze and address the depth regularization problem for 3D Gaussian Splatting under coarse depth cues. 
We hope this paper can inspire more ideas for optimizing radiance fields in under-constrained situations.


\vspace{-0.1cm}
\section{Related Work}
\vspace{-0.1cm}

\vspace{0pt}\noindent\textbf{Radiance Fields for Novel View Synthesis.} Novel view synthesis aims to generate unseen views of the same object or scene from a set of given images \cite{zhou2016view, avidan1997novel}. Neural Radiance Fields (NeRF) \cite{mildenhall2021nerf}
uses a large MLP to represent 3D scenes and renders via volume rendering. However, its speed is slow both in training and inference. 
The following improvements mainly pursue either higher quality \cite{barron2021mip, barron2022mip360} or efficiency \cite{chen2022tensorf, muller2022instant, sun2022direct, yu2021plenoxels, yu2021plenoctrees, liu2020nsvf, hu2023tri}, but hard to achieve both.
The most recent unstructured radiance fields \cite{chen2023neurbf, xu2022point, kerbl20233d} utilize a set of primitives to represent scenes.
Among them, 3D Gaussian Splatting \cite{kerbl20233d} represents radiance fields by a set of anisotropic 3D Gaussians and renders with a differentiable splatting. This approach achieves great success in fast and high-quality reconstruction for complex real scenes. 
While this method excels with dense input views and has achieved success in various 3D tasks \cite{luiten2023dynamic, wu20234d, tang2023dreamgaussian}, its reconstruction with sparse view inputs remains an open problem. Also, issues such as how to apply additional constraints for improvement are still unsolved and worthy of discussion.

\vspace{2pt}\noindent\textbf{Few-shot Novel View Synthesis.}
Few-shot novel view synthesis aims to generate novel views from only a set of sparse input views. Many works address the problem by introducing regularization strategies specified for NeRF \cite{yang2023freenerf, niemeyer2022regnerf, kim2022infonerf, deng2022dsnerf}. Some pre-trained methods aim to design a generative model and train it on large datasets \cite{chen2021mvsnerf, yu2021pixelnerf, cong2023enhancing, zhou2023sparsefusion, kulhanek2022viewformer}, while others \cite{wynn2023diffusionerf, jain2021putting} take pre-trained models as a type of loss to regularize the training process with well-learned knowledge. Depth distilling \cite{deng2022dsnerf, roessle2022dense, song2023darf, wang2023sparsenerf} is also a powerful technique for sparse-view neural fields. 
However, limited by their powerful but slow backbones or the complex pre-trained models, most of these methods are costly in both training and inference. Although some methods \cite{song2023darf, sun2023vgos, wynn2023diffusionerf} have improved inference efficiency via grid-based backbones, they also suffer from trade-offs like higher training costs or lower quality. 
More recently, some work \cite{liu2023zero, sargent2023zeronvs, qian2023magic123} enable zero-shot novel view synthesis with even one input by diffusion model priors, but can hardly handle complex scenes and with lower efficiency. 

\vspace{2pt}\noindent\textbf{Depth Supervision in Sparse-view Neural Fields.} 
As a classic cue in many 3D vision tasks \cite{wang2024contrastive, wang2024robust, wang2022uncertainty, zhang2022revisiting, wang2021multi, zhou2023Adaptive}, depth information has been widely used to supervise sparse-view neural fields. The first group \cite{deng2022dsnerf, roessle2022dense} is to extract accurate but sparse depth values from reliable point clouds, and the second \cite{yu2022monosdf, hu2023consistentnerf, song2023darf, uy2023scade, wang2023sparsenerf} distills depth knowledge from current powerful monocular depth estimators \cite{Ranftl2022midas, Ranftl2021dpt}. 
Considering point clouds are sparse and not available in many sparse-view cases, monocular depth shows its advantage in density and robustness for our tasks. To tackle the scale ambiguity and error of monocular depths, some previous works and concurrent sparse-view 3DGS methods have introduced various scale-invariant losses \cite{yu2022monosdf, deng2023nerdi, song2023darf, xiong2023sparsegs, zhu2023fsgs} including depth ranking loss \cite{xu2023neurallift, wang2023sparsenerf}, however, all of which are not optimal for us. Firstly, flexible Gaussians are more sensitive to wrong depth cues, requiring extra designs for regularization. Also, these losses align the depth to a certain fixed global scale, which may ignore minor local depth changes. This overlook can lead to a noisy primitive distribution, particularly in regions with intricate textures. 
Besides, we notice an HDN loss \cite{zhang2022hierarchical} that can preserve details in monocular depth estimation. Nevertheless, it is also unsuitable as its reliance on multi-scale patches would bring long-distance errors and compromise geometric accuracy.

\vspace{-0.2cm}
\section{Method}
\vspace{-0.1cm}

\begin{figure*}[t]
    \setlength{\abovecaptionskip}{14pt}

    \centering
    \includegraphics[width=1\linewidth]{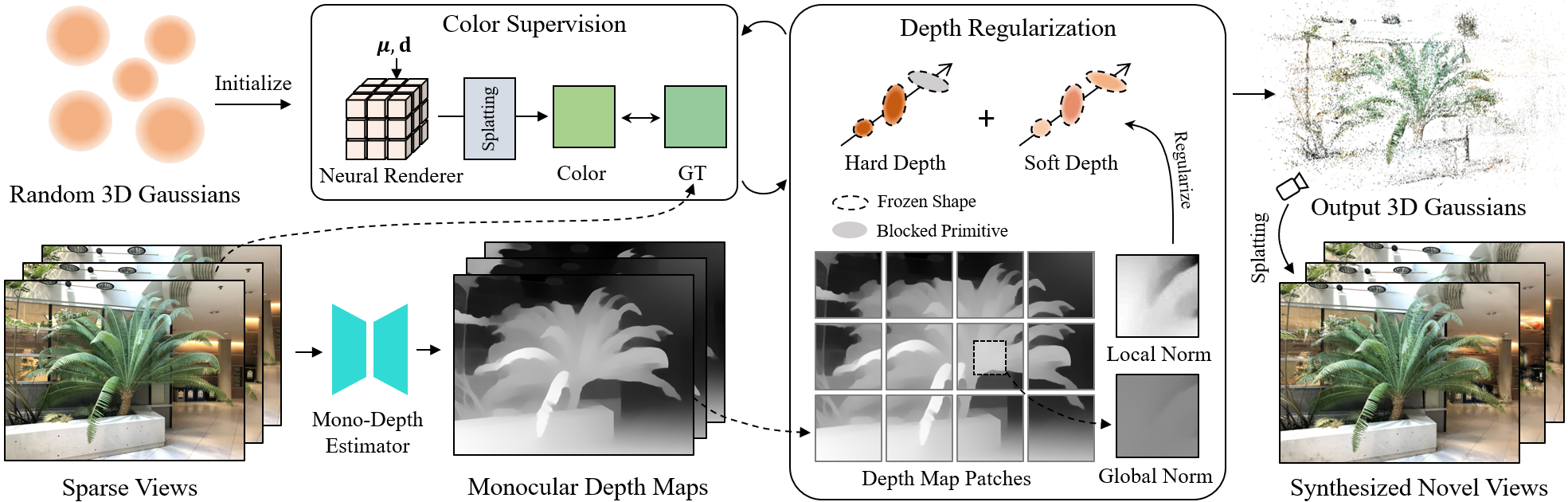}
    \caption{\textbf{The Framework of DNGaussian.} Our framework starts from a random initialization and consists of a Color Supervision module and a Depth Regularization module. The optimization process of Color Supervision mainly inherits from 3D Gaussian Splatting \cite{kerbl20233d} except for a Neural Color Renderer.  In the depth regularization, we render a Hard Depth and a Soft Depth for the input view, and separately calculate the losses of the pre-generated monocular depth map with the proposed Global-Local Depth Normalization. Finally, the output Gaussian field enables efficient and high-quality novel view synthesis. }
    \label{fig:main}
    \vspace{-.3cm}
\end{figure*}

\subsection{Preliminary for 3D Gaussian Splatting} 
\vspace{-0.1cm}

\vspace{0pt}\noindent\textbf{Representation. } 
3D Gaussian splatting \cite{kerbl20233d} represents 3D information with a set of 3D Gaussians. It computes pixel-wise color $\mathcal{C}$ with a set of 3D Gaussian primitives $\theta$, view pose $P$, and the camera parameter involving the center $o$.

Specifically, a Gaussian primitive can be described with a center $\mu \in \mathbb{R}^3$, a scaling factor $s \in \mathbb{R}^3$, and a rotation quaternion $q \in \mathbb{R}^4$. The basis function of the $i$-th primitive $\mathcal{G}_i$ is in the form of:
\begin{equation}
\setlength{\abovedisplayskip}{4pt}
\setlength{\belowdisplayskip}{4pt}
    \mathcal{G}_i(x) = e^{-\frac{1}{2}(x-\mu_i)^T\Sigma_i^{-1}(x-\mu_i)},
\end{equation}
where the covariance matrix $\Sigma$ can be calculated from the scale $s$ and rotation $q$.
For rendering purposes, the Gaussian primitive also retains an opacity value $\alpha \in \mathbb{R}$ and a $K$-dimensional color feature $f \in \mathbb{R}^K$. Then $\theta_i = \{\mu_i, s_i, q_i, \alpha_i, f_i\}$ is the parameters for the $i$-th Gaussian.

\vspace{2pt}\noindent\textbf{Rendering. } 
 3D Gaussian Splatting utilizes a point-based rendering to compute the color $\mathcal{C}$ of pixel $x_p$ by blending $N$ ordered Gaussians overlapping the pixel:
\begin{equation}
    \setlength{\abovedisplayskip}{2pt}
    \setlength{\belowdisplayskip}{2pt}
    \mathcal{C}(x_p) = \sum_{i \in N}{c_i\widetilde{\alpha}_i\prod_{j=1}^{i-1}(1-\widetilde{\alpha}_j)},
\end{equation}
where $c_i$ is the decoded color of feature $f$. 

Different from NeRF's ray sampling strategy, the involved $N$ Gaussians are gathered by a well-optimized rasterizer according to $x_p$, the camera parameter, the view pose $P$, and a set of pre-defined roles. And the rendering opacity $\widetilde{\alpha}$ of $N$ primitives are calculated by $\alpha$ and their projected 2D Gaussians $\mathcal{G}^{proj}$ on image plane :
\begin{equation}
\setlength{\abovedisplayskip}{1pt}
\setlength{\belowdisplayskip}{3pt}
    \widetilde{\alpha}_i = \alpha_i \mathcal{G}^{proj}_i(x_p).
\end{equation}
Then, similar to NeRF, we can represent the pixel-wise depth $\mathcal{D}$ with the distance to the camera center $o$:
\begin{equation}
\setlength{\abovedisplayskip}{2pt}
\setlength{\belowdisplayskip}{2pt}
    \mathcal{D}(x_p) = \sum_{i \in N}{||\mu_i - o||_2} \times \widetilde{\alpha}_i \prod_{j=1}^{i-1}(1-\widetilde{\alpha}_j).
\end{equation}

\noindent\textbf{Optimzation. } 3D Gaussian Splatting optimizes the parameters $\theta$ for all Gaussians through gradient descent under color supervision. During the optimization process, it identifies and duplicates the most active primitives to better represent intricate textures, simultaneously removing redundant primitives. In this work, we inherit these optimization strategies for color supervision.

\vspace{2pt}\noindent\textbf{Initialization.} To start from a better geometry, the method suggests utilizing the point cloud from COLMAP \cite{schoenberger2016mvs, schoenberger2016sfm} or other SfMs to initialize the Gaussians. Instead, considering the instability of point clouds in sparse-view situations, we initialize our method with a random set of Gaussians.

\vspace{-0.1cm}
\subsection{Depth Regularization for Gaussians \label{sec:reg}}
\vspace{-.1cm}
Despite sharing a similar depth computation, existing depth regularization for NeRFs cannot transfer to 3D Gaussian radiance fields due to the huge differences. First, a target conflict between color and depth would occur in the extra parameters. Also, previous regularization for the continuous NeRF only focuses on density, for which it can hardly work well on the discrete and flexible Gaussian primitives.

\vspace{2pt}\noindent\textbf{Shape Freezing.}
3D Gaussian radiance fields possess four optimizable parameters $\{\mu, s, q, \alpha\}$ that can directly influence the depth, which is more complex than NeRF. Since the mono-depth is much smoother and easier to fit than color, apply an all-parameter depth regularization on the entire model, which is widely used in previous sparse-view neural fields \cite{yu2022monosdf, wang2023sparsenerf, hu2023consistentnerf, deng2023nerdi, xu2023neurallift}, would lead the shape parameters to overfit the target depth map and cause blurry appearance. 
Thus, these parameters must be treated differently.
Since the scene geometry is mainly represented by the position distribution of Gaussian primitives, we regard the \textit{ center $\mu$} and \textit{opacity $\alpha$} as the most important parameters to regularize, for they separately stand for the position itself and the occupancy of a position. 
Furthermore, to reduce the negative influence for color reconstruction, we \textit{freeze the scaling $s$ and rotation $q$} in the depth regularization.

\vspace{2pt}\noindent\textbf{Hard Depth Regularization.}
To achieve the spatial reshaping of the Gaussian fields, we first propose a Hard Depth Regularization that encourages the movement of the nearest Gaussians, which are expected to compose surfaces but often cause noises and artifacts. Considering the predicted depth is rendered with the mixture of multiple Gaussians and reweighted by the cumulative product $\widetilde{\alpha}$, we manually apply a large opacity value $\tau$ to all Gaussians. Then, we render a ``hard depth" that mainly consists of the nearest Gaussians on the ray shot from camera center $o$ and across the pixel $x_p$:
\begin{equation}
\setlength{\abovedisplayskip}{2pt}
\setlength{\belowdisplayskip}{2pt}
    \mathcal{D}_{hard}(x_p) = \sum_{i \in N}{\tau (1-\tau)^{i-1}\mathcal{G}^{proj}_i(x_p) ||\mu_i - o||_2 }.
\end{equation}
Since now only the center $\mu$ is in optimization, Gaussians at wrong positions cannot avoid being regularized by decreasing their opacity or changing shapes, and thus their centers $\mu$ move. The regularization is implemented by a similarity loss at the target image area $\mathcal{P}$ to encourage the hard depth $\mathcal{D}_{hard}$ close to the monocular depth $\widetilde{\mathcal{D}}$:
\begin{equation}
    \mathcal{R}_{hard}(\mathcal{P}) = \mathcal{L}_{similar}(\mathcal{D}_{hard}(\mathcal{P}), \widetilde{\mathcal{D}}(\mathcal{P})).
\end{equation}

\noindent\textbf{Soft Depth Regularization.}
Only regularizing on ``hard depth" is insufficient due to the absence of opacity optimization. We also expect to ensure the accuracy of the real rendered ``soft depth", otherwise, the surface may become semitransparent and cause hollowness. From this perspective, we additionally \textit{freeze the Gaussian center $\mu$} (denoted by $\check{\mu}$) to avoid negative influence caused by the center moving, and propose Soft Depth Regularization for the tuning of the opacity $\alpha$:
\begin{equation}
\begin{aligned}
     \mathcal{D}_{soft}(x_p) = \sum_{i \in N}{||\check{\mu_i} - o||_2} \times \widetilde{\alpha}_i \prod_{j=1}^{i-1}(1-\widetilde{\alpha}_j), \\ 
   \mathcal{R}_{soft}(\mathcal{P}) = \mathcal{L}_{similar}(\mathcal{D}_{soft}(\mathcal{P}), \widetilde{\mathcal{D}}(\mathcal{P})).
\end{aligned}
\end{equation}

With both the Hard and Soft Depth Regularization, we constrain the nearest Gaussians to stay in a suitable position with high opacity, therefore composing complete surfaces.

\vspace{-0.1cm}
\subsection{Global-Local Depth Normalization}
\vspace{-0.1cm}

Previous depth-supervised neural fields usually build the depth loss on the source scales of the depth maps\cite{yu2022monosdf, song2023darf, deng2023nerdi, hu2023consistentnerf, wang2023sparsenerf}. 
This type of alignment measures all losses via a fixed scale based on the statistics of a large area. As a result, it might lead to the overlooking of small errors, particularly when dealing with multiple objectives such as color reconstruction or a wide range of depth variance.
This overlook may matter not much in previous NeRF-based works, but can raise heavier problems in the Gaussian radiance fields. 

\begin{figure}[t]
    \centering
    \setlength{\abovecaptionskip}{5pt}
    \includegraphics[width=\linewidth]{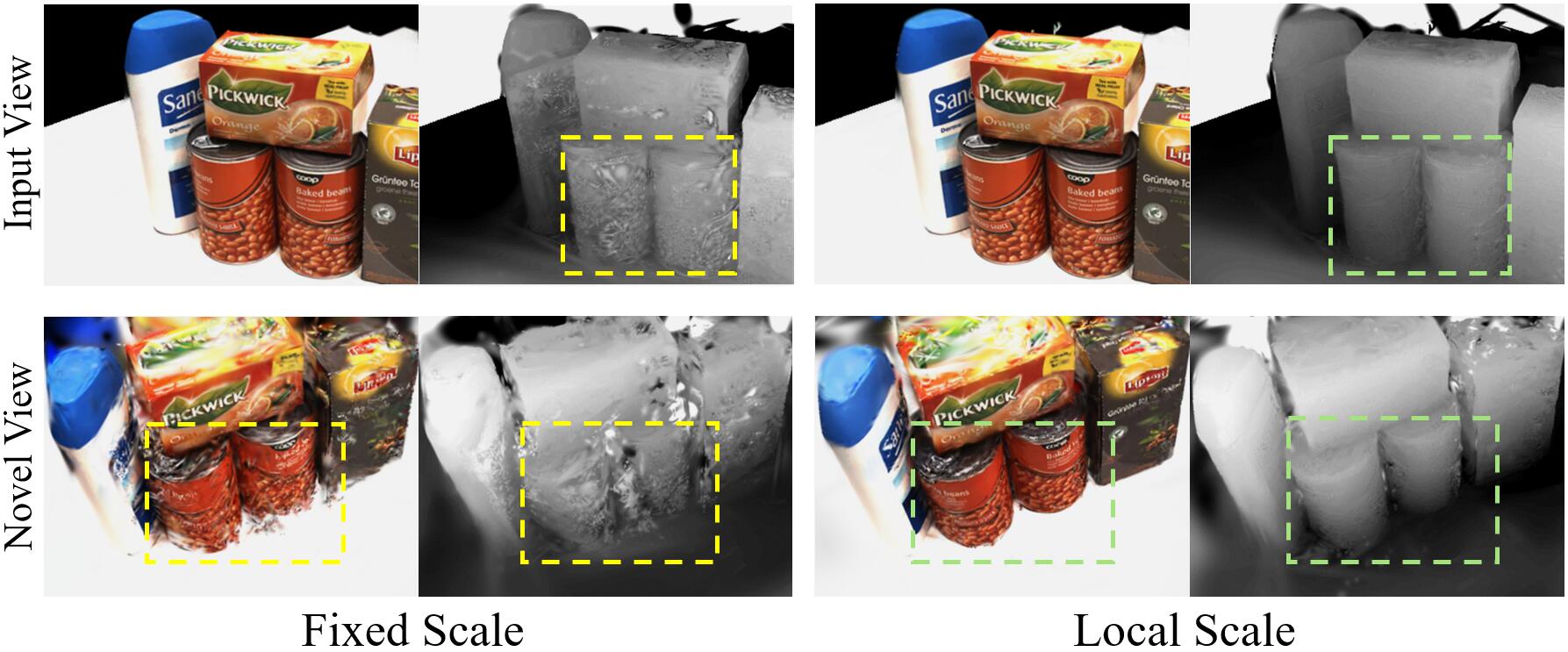}
    \caption{A fixed global scale pays little attention to the small depth errors even under L1 loss, which leads to noisy primitives and causes failures in novel view (yellow box). Our Global-Local Depth Normalization refocuses on small errors via local scale and helps reconstruct a more accurate appearance (green box).}
    \label{fig:scale}
    \vspace{-.5cm}
\end{figure}

In the Gaussian radiance fields, correcting small depth errors is more challenging because it primarily relies on the movement of Gaussian primitives, a process that happens with a minor learning rate.
Also, if the primitives have not been corrected in position during depth regularization, they will become float noises and cause failures, especially in regions with detailed appearance where gathering numerous primitives, as shown in Figure \ref{fig:scale}. 

\vspace{2pt}\noindent\textbf{Local Depth Normalization.}
To solve this problem, we make the loss function refocus on small errors by introducing a patch-wise local normalization. Specifically, we cut a whole depth map into small patches and normalize the patch $\mathcal{P}$ of predicted depth and monocular depth with the mean value of $0$ and standard deviation of near to $1$:
\begin{equation}
\setlength{\abovedisplayskip}{5pt}
\setlength{\belowdisplayskip}{5pt}
    \mathcal{D}^{LN}(x) = \frac{\mathcal{D}(x) - \text{mean}(\mathcal{D}(\mathcal{P}))}{\text{std}(\mathcal{D}(\mathcal{P})) + \epsilon}, \quad  \mathrm{s.t.}\ \ x \in \mathcal{P},
\end{equation}
where $\epsilon$ is a value for numerical stability. Since then, all patches have been normalized on a local scale and the loss can be calculated inside. Later, we apply the Local Depth Normalization to the Hard and Soft Depth Regularization to help with geometry reshaping. 

\vspace{2pt}\noindent\textbf{Global Depth Normalization.}
In contrast to focusing on small local losses, we also need a global view to learn an overall shape. To fill the lack of global scale, we further add a Global Depth Normalization in the depth regularization. This makes the depth loss aware of the global scale while preserving local relevance. Similar to the local one, we apply a patch-wise normalization to free the depth from the source scale and focus on local changes. The only difference is here we use a global standard deviation of the whole image depth $\mathcal{D}_\mathcal{I}$ of image $\mathcal{I}$ to replace that of the patch:
 \begin{equation}
 \setlength{\abovedisplayskip}{3pt}
\setlength{\belowdisplayskip}{1pt}
 \begin{aligned}
     \mathcal{D}^{GN}(x) = \frac{\mathcal{D}(x) - \text{mean}(\mathcal{D}(\mathcal{P}))}{\text{std}(\mathcal{D}_I)},& \\ 
     \mathrm{s.t.}\ \ x \in \mathcal{P}, \ & \mathcal{P} \subseteq \mathcal{I}.
 \end{aligned}
 \end{equation}

In addition, our patch-wise normalization can also avoid long-distance errors in the monocular depth by driving the learning of locally relative depth, which serves a similar effect as depth rank distillation \cite{wang2023sparsenerf, xu2023neurallift}. But differently, for geometry reshaping purposes, we also encourage the model to learn the absolute depth change rather than ignoring it.

\vspace{-0.1cm}
\subsection{Training Details}
\vspace{-0.1cm}
\noindent\textbf{Loss Function} 
The loss function consists of three parts: color reconstruction loss $\mathcal{L}_{color}$, hard depth regularization $\mathcal{R}_{hard}$ and soft depth regularization $\mathcal{R}_{soft}$. 
Following 3D Gaussian Splatting, the color reconstruction loss is a combination of L1 reconstruction loss and a D-SSIM term of the rendering image $\hat{\mathcal{I}}$ and ground-truth $\mathcal{I}$:
\begin{equation}
 \setlength{\abovedisplayskip}{6pt}
\setlength{\belowdisplayskip}{6pt}
    \mathcal{L}_{color} = \mathcal{L}_1(\hat{\mathcal{I}}, \mathcal{I}) + \lambda \mathcal{L}_{\mathrm{D-SSIM}}(\hat{\mathcal{I}}, \mathcal{I}) .
\end{equation}
The depth regularization $\mathcal{R}_{hard}$ and $\mathcal{R}_{soft}$ all include a local and a global term separately from our two kinds of depth normalization. We take the L2 loss to measure the similarity. Both of the regularizations are in the form of:
\begin{equation}
 \setlength{\abovedisplayskip}{6pt}
\setlength{\belowdisplayskip}{6pt}
    \mathcal{R}_{T} = \mathcal{L}_{2}(\mathcal{D}_{T}^{GN}, \widetilde{\mathcal{D}}^{GN}) + \gamma \mathcal{L}_{2}(\mathcal{D}_{T}^{LN}, \widetilde{\mathcal{D}}^{LN}) ,
\end{equation}
where $T$ stands for $hard$ or $soft$. In practice, we reserve an error tolerance for the L2 loss to relax the constraint.
The full loss function is formulated by:
\begin{equation}
 \setlength{\abovedisplayskip}{4pt}
\setlength{\belowdisplayskip}{4pt}
    \mathcal{L} = \mathcal{L}_{color} + \mathcal{R}_{hard} + \mathcal{R}_{soft} .
\end{equation}

\begin{table*}[!t]
\setlength{\abovecaptionskip}{4pt}
\resizebox{1\linewidth}{!}{
\setlength{\tabcolsep}{3.2 mm}
\begin{tabular}{l|c|cccccccc}
\toprule
             &                                                                                                      & \multicolumn{4}{c}{LLFF}                                                                                                                           & \multicolumn{4}{c}{DTU}                                                                                                       \\
             & \multirow{-2}{*}{Setting}                                                                            & PSNR $\uparrow$               & LPIPS $\downarrow$             & SSIM $\uparrow$                 & \multicolumn{1}{c|}{AVGE $\downarrow$}         & PSNR $\uparrow$              & LPIPS $\downarrow$             & SSIM $\uparrow$             & AVGE $\downarrow$             \\ \midrule
SRF \cite{chibane2021stereo} &                                                                                                      & 12.34                         & 0.591                         & 0.250                         & \multicolumn{1}{c|}{0.313}                         & 15.32                         & 0.304                         & 0.671                         & 0.171                         \\
PixelNeRF \cite{yu2021pixelnerf}  &                                                                                                      & 7.93                          & 0.682                         & 0.272                         & \multicolumn{1}{c|}{0.461}                         & 16.82                         & 0.270                         & 0.695                         & 0.147                         \\
MVSNeRF \cite{chen2021mvsnerf} & \multirow{-3}{*}{Trained on DTU}                                                                     & 17.25                         & 0.356                         & 0.557                         & \multicolumn{1}{c|}{0.171}                         & 18.63                         & 0.197                         & \cellcolor[HTML]{FFFFD4}0.769            & 0.113                         \\ \midrule
SRF ft \cite{chibane2021stereo}  &                                                                                                      & 17.07                         & 0.529                         & 0.436                         & \multicolumn{1}{c|}{0.203}                         & 15.68                         & 0.281                         & 0.698                         & 0.162                         \\
PixelNeRF ft \cite{yu2021pixelnerf} &                                                                                                      & 16.17                         & 0.512                         & 0.438                         & \multicolumn{1}{c|}{0.217}                         & 18.95                         & 0.269                         & 0.710                         & 0.125                         \\
MVSNeRF ft \cite{chen2021mvsnerf} & \multirow{-3}{*}{\begin{tabular}[c]{@{}c@{}}Trained on DTU \\ Fine-tuned per Scene\end{tabular}} & 17.88                         & \cellcolor[HTML]{FFFFD4}0.327 & 0.584                         & \multicolumn{1}{c|}{0.157}                         & 18.54                         & 0.197                         & \cellcolor[HTML]{FFFFD4}0.769 & 0.113                         \\ \midrule
Mip-NeRF \cite{barron2021mip} &                                                                                                      & 14.62                         & 0.495                         & 0.351                         & \multicolumn{1}{c|}{0.246}                         & 8.68                          & 0.353                         & 0.571                         & 0.323                         \\
DietNeRF \cite{jain2021putting}  &                                                                                                      & 14.94                         & 0.496                         & 0.370                         & \multicolumn{1}{c|}{0.240}                         & 11.85                         & 0.314                         & 0.633                         & 0.243                         \\
RegNeRF \cite{niemeyer2022regnerf} &                                                                                                      & 19.08                         & 0.336                         & 0.587                         & \multicolumn{1}{c|}{0.149}                         & 18.89             & \cellcolor[HTML]{FFFFD4}0.190 & 0.745                         & 0.112                         \\
FreeNeRF \cite{yang2023freenerf} &                                                                                                      & \cellcolor[HTML]{FFE4CF}19.63 & \cellcolor[HTML]{FFE4CF}0.308 & \cellcolor[HTML]{FFE4CF}0.612 & \multicolumn{1}{c|}{\cellcolor[HTML]{FFFFD4}0.134} & \cellcolor[HTML]{FFCCC9}19.92                     & \cellcolor[HTML]{FFE4CF}0.182 & \cellcolor[HTML]{FFE4CF}0.787 & \cellcolor[HTML]{FFCCC9}0.098 \\
SparseNeRF \cite{wang2023sparsenerf} & \multirow{-5}{*}{Optimized per Scene}                                                                & \cellcolor[HTML]{FFCCC9}19.86 & 0.328                         & \cellcolor[HTML]{FFCCC9}0.624 & \multicolumn{1}{c|}{\cellcolor[HTML]{FFCCC9}0.127} & \cellcolor[HTML]{FFE4CF}19.55                     & 0.201                         & \cellcolor[HTML]{FFFFD4}0.769 & \cellcolor[HTML]{FFE4CF}0.102 \\ \midrule
3DGS \cite{kerbl20233d}    &                                                                                          &  15.52     & 0.405             & 0.408                 & \multicolumn{1}{c|}{0.209}                              &  10.99                                        &       0.313                   &   0.585             &  0.252                     \\
3DGS\dag                        &                                                                                          &  16.46     & 0.401             & 0.440                 & \multicolumn{1}{c|}{0.192}                              & 14.74                         &  0.249                        &  0.672                   & 0.169                       \\

\textbf{DNGaussian (Ours)} & \multirow{-3}{*}{Optimized per Scene}                                                                & \cellcolor[HTML]{FFFFD4}19.12 & \cellcolor[HTML]{FFCCC9}0.294 & \cellcolor[HTML]{FFFFD4}0.591 & \multicolumn{1}{c|}{\cellcolor[HTML]{FFE4CF}0.132} & \cellcolor[HTML]{FFFFD4}18.91 & \cellcolor[HTML]{FFCCC9}0.176 & \cellcolor[HTML]{FFCCC9}0.790 & \cellcolor[HTML]{FFE4CF}0.102 \\ \bottomrule
\multicolumn{5}{l}{\small\dag \ with the same hyperparameters and the neural color renderer as DNGaussian . }
\end{tabular}
}
\caption{\textbf{Quantitative Comparison on LLFF and DTU for 3 input views.} The best, second-best, and third-best entries are marked in red, orange, and yellow, respectively. Notably, the Gaussian-based methods directly show the background color on the meaningless invisible places, which would cause lower metrics, especially in PSNR.}
\label{tab:llffdtu}
\vspace{-.4cm}
\end{table*}

\begin{table}[t]
\setlength{\abovecaptionskip}{4pt}

\resizebox{1\linewidth}{!}{
\setlength{\tabcolsep}{5 mm}
\centering
\begin{tabular}{l|ccc}
\toprule
Method            & PSNR $\uparrow$                & SSIM $\uparrow$                  & LPIPS $\downarrow$               \\ \midrule
NeRF \cite{mildenhall2021nerf}  & 14.934                         & 0.687                         & 0.318                         \\
NeRF (Simplified) \cite{jain2021putting}  & 20.092                         & 0.822                         & 0.179                         \\
DietNeRF \cite{jain2021putting}   & 23.147                         & 0.866                         & 0.109                         \\
DietNeRF + ft \cite{jain2021putting} & \cellcolor[HTML]{FFFFD4}23.591   & \cellcolor[HTML]{FFFFD4}0.874 & \cellcolor[HTML]{FFE4CF}0.097 \\
FreeNeRF \cite{yang2023freenerf} & \cellcolor[HTML]{FFE4CF}24.259 & \cellcolor[HTML]{FFE4CF}0.883 & \cellcolor[HTML]{FFFFD4}0.098 \\ 
SparseNeRF \cite{wang2023sparsenerf} & 22.410                         & 0.861                         & 0.119                         \\ \midrule
3DGS \cite{kerbl20233d} & 22.226                         & 0.858                         & 0.114                         \\
\textbf{DNGaussian (Ours)}        & \cellcolor[HTML]{FFCCC9}24.305 & \cellcolor[HTML]{FFCCC9}0.886 & \cellcolor[HTML]{FFCCC9}0.088 \\ \bottomrule
\end{tabular}
}
\caption{\textbf{Quantitative Comparison on Blender for 8 input views.} The best, second-best, and third-best entries are marked in red, orange, and yellow, respectively.}
\label{tab:blender}
\vspace{-.6cm}
\end{table}

\vspace{2pt}\noindent\textbf{Neural Color Renderer.}
3D Gaussian Splatting stores color via spherical harmonics, however, it is easy to overfit with only sparse views. To relieve this problem, we take a grid encoder and an MLP as the Neural Color Renderer to predict color for each primitive (Figure \ref{fig:main}). During inference, we store the intermediate result and only calculate the last MLP layers to merge view direction for acceleration.  


\vspace{-.1cm}
\section{Experiments}
\vspace{-.1cm}
\subsection{Setups}
\vspace{-.1cm}

\noindent\textbf{Datasets.}
we conduct our experiment on three datasets: the NeRF Blender Synthetic dataset (Blender) 
 \cite{mildenhall2021nerf}, the DTU dataset \cite{jensen2014large}, and the LLFF dataset \cite{mildenhall2019local}.
We follow the setting used in previous works \cite{niemeyer2022regnerf, yang2023freenerf, wang2023sparsenerf} with the same split of DTU and LLFF to train the model on 3 views and test on another set of images. To erase the noises in the background and focus on the target object, we apply the same object masks as previous works \cite{niemeyer2022regnerf} for DTU at evaluation. For Blender, we follow DietNeRF \cite{jain2021putting} and FreeNeRF \cite{yang2023freenerf} to train with the same 8 views and test on 25 unseen images. Aligned with the baselines, downsampling rates of $8$, $4$, and $2$ are applied to LLFF, DTU, and Blender. Following previous sparse-view settings, the camera poses are assumed to be known via calibration or other ways.

\vspace{2pt}\noindent\textbf{Evaluation Metrics.}
We report PSNR, SSIM \cite{wang2004ssim}, and LPIPS \cite{zhang2018lpips} scores to evaluate the reconstruction performance quantitatively. Also, an Average Error (AVGE) \cite{niemeyer2022regnerf} is reported by the geometric mean of $\text{MSE} = 10^{-\text{PSNR}/10}$, $\sqrt{1 - \text{SSIM}}$, and LPIPS.

\vspace{2pt}\noindent\textbf{Baselines.}
Following the previous sparse-view neural fields \cite{niemeyer2022regnerf, jain2021putting, yang2023freenerf, wang2023sparsenerf}, We take current SOTA methods SRF \cite{chibane2021stereo}, PixelNeRF \cite{yu2021pixelnerf}, MVSNeRF \cite{chen2021mvsnerf}, Mip-NeRF \cite{barron2021mip}, DietNeRF \cite{jain2021putting}, RegNeRF \cite{niemeyer2022regnerf}, FreeNeRF \cite{yang2023freenerf}, and SparseNeRF \cite{wang2023sparsenerf} as our baselines. For most NeRF-based methods, we directly report their best quantitative results in corresponding published papers for comparisons. The results of raw 3D Gaussian Splatting (3DGS) \cite{kerbl20233d} are also reported.

\vspace{2pt}\noindent\textbf{Implementations.} 
We build our models on the official PyTorch 3D Gaussian Splatting codebase. We train the model with $6,000$ iterations for all datasets, and the soft depth regularization is applied after $1,000$ iterations for stability. We set $\gamma = 0.1, \tau=0.95$ in loss functions for all experiments. The neural renderer consists of a hash encoder \cite{muller2022instant} with $16$ levels in a resolution range of $16$ to $512$ and a max size of $2^{19}$, and a $5$ layer MLP with the hidden dim of $64$. We use DPT \cite{Ranftl2021dpt} to predict monocular depth maps for all input views. The models of 3DGS and DNGaussian are randomly initialized with a uniform distribution.


\vspace{-0.1cm}
\subsection{Comparison}
\vspace{-0.1cm}

\begin{figure*}[!t]\
    \setlength{\abovecaptionskip}{4pt}
    \centering
    \includegraphics[width=1\textwidth]{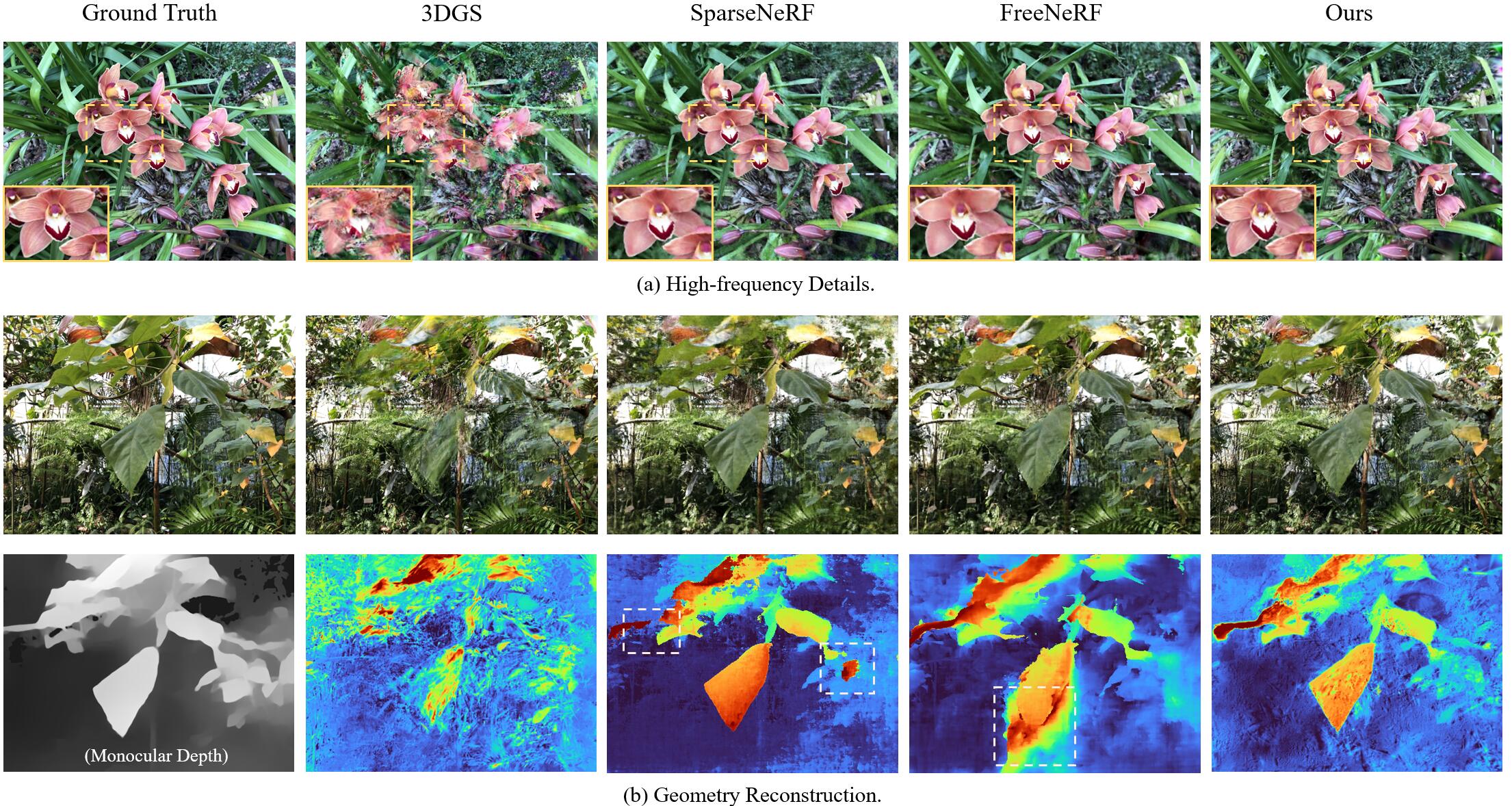}
    \caption{\textbf{Qualitative Comparison on LLFF.} 3DGS \cite{kerbl20233d} fails to synthesize accurate novel views under sparse inputs. The rendering views from FreeNeRF \cite{yang2023freenerf} and SparseNeRF \cite{wang2023sparsenerf} are both smooth but with too many details lost. FreeNeRF further learns a wrong geometry in complete scenes. Our method learns more complete foreground geometry and renders high-quality novel views with fine details.}
    \label{fig:llff}
    \vspace{-.5cm}
\end{figure*}

\noindent\textbf{LLFF.}
The qualitative results and visualizations on the LLFF dataset from 3 input views are reported in Table \ref{tab:llffdtu} and Figure \ref{fig:llff}. Notably, since the NeRF-based baselines would interpolate colors to those invisible areas from input views while the discrete Gaussian radiance fields directly expose the black background on these empty spaces, the 3DGS-based methods natively have a weakness in the reconstruction metrics from these meaningless invisible areas. Despite that, our approach still outperforms all baselines in the LPIPS score, and achieves comparable PSNR, SSIM, and Average Error to the best methods. From both the quantitative and qualitative results, we can see that our DNGaussian predicts more fine details and precise geometry. FreeNeRF tends to synthesize smooth views that lack high-frequency details, also the geometry is not as accurate as the depth-supervised SparseNeRF and Our DNGaussian. Although regularized by same depth maps, SparseNeRF performs more weak in details and geometry completeness. DNGaussian also has huge improvements in both image geometry quality compared to the well-tuned 3DGS.

\begin{figure}[t]
    \setlength{\abovecaptionskip}{5pt}

    \centering
    \includegraphics[width=\linewidth]{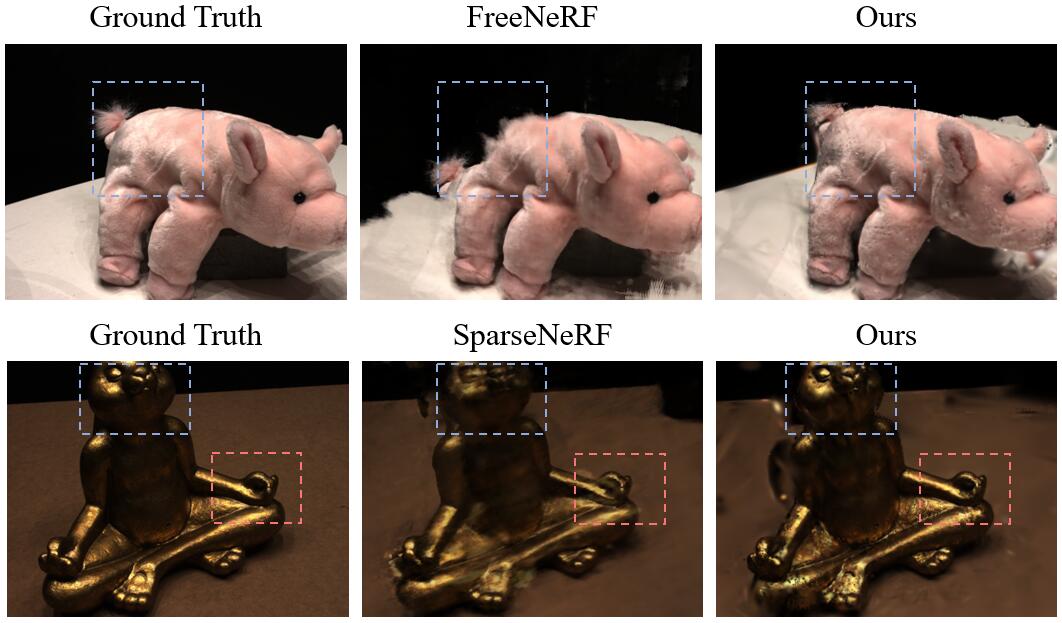}
    \caption{\textbf{Qualitative Comparison on DTU.} Our method excels both in geometry and rendering qualities in difficult areas.}
    \label{fig:dtu}
    \vspace{-.7cm}
\end{figure}

\vspace{2pt}\noindent\textbf{DTU.}
The quantitative results on the DTU 3-view setting reported in Table \ref{tab:llffdtu} show that our method achieves the best in LPIPS and SSIM, and the second best in Average Error. However, we got a lower score in PSNR, which is mainly due to scale variance and the noise occlusion coming from the textureless board and background in the scene. In the qualitative examples in Figure \ref{fig:dtu}, It can be observed that our method can learn a more correct and complete geometry compared with both FreeNeRF and the depth-supervised SparseNeRF, and produces high-quality details even on difficult plush and reflective areas.

\begin{figure}[t]
    \setlength{\abovecaptionskip}{5pt}

    \centering
    \includegraphics[width=1\linewidth]{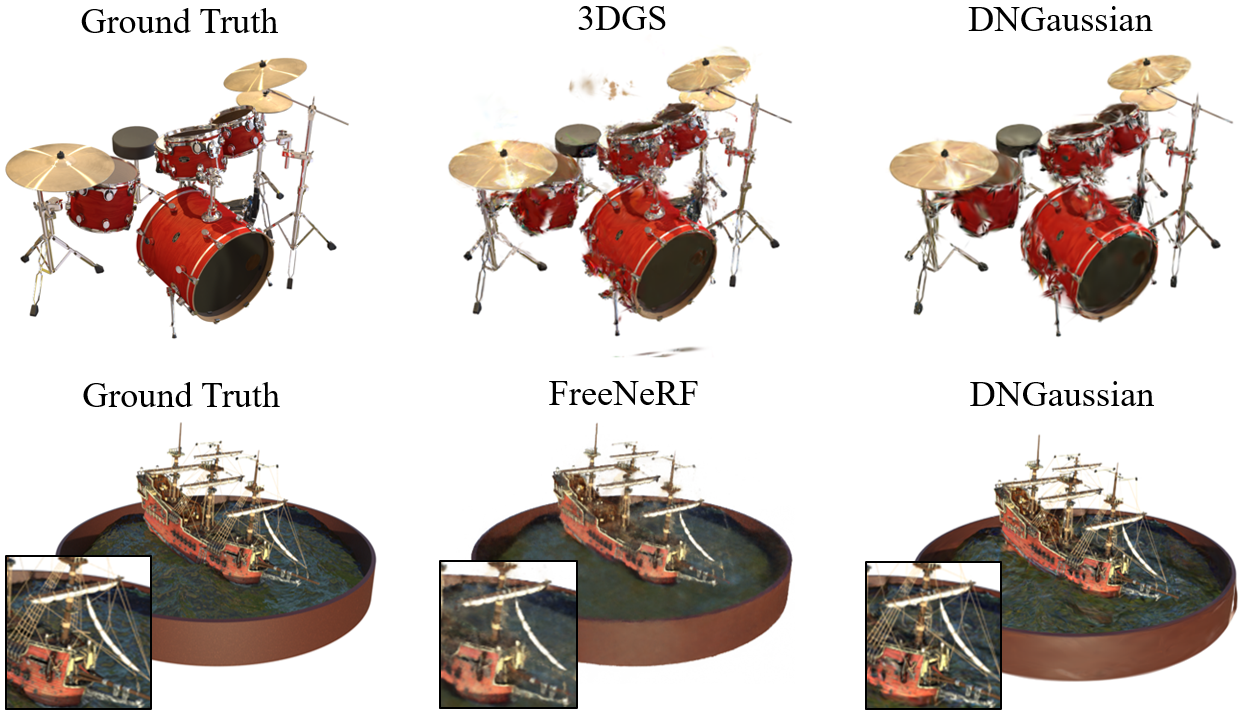}
    \caption{\textbf{Qualitative Comparison on Blender.} The results demonstrate the strong fitting ability from surrounding views and reconstruct detailed and complex scenes.}
    \label{fig:blender}
    \vspace{-.6cm}
\end{figure}

\vspace{2pt}\noindent\textbf{Blender.}
To test the fitting ability from surrounding views, we make an evaluation of the Blender dataset under 8 input views. The scores are reported in Table \ref{tab:blender}, in which some data come from FreeNeRF \cite{yang2023freenerf} and DietNeRF \cite{jain2021putting}. Our method has got the best scores in all PSNR, SSIM and LPIPS. From the qualitative results in Figure \ref{fig:blender}, it can be seen that our method synthesizes views with correct geometry and fewer floaters compared to the vanilla 3DGS, and has a better performance in detail compared to the second-best FreeNeRF. The results demonstrate that DNGaussian can not only handle looking-forward scenes like LLFF and DTU, but also a whole reconstruction of complex objects with transparent and reflective materials.

\vspace{2pt}\noindent\textbf{Efficiency.}
We further conduct an efficiency study on the LLFF 3-view setting with RTX 3090 Ti GPUs to explore the performance of current SOTA baselines \cite{wang2023sparsenerf, yang2023freenerf} with limited GPU memories of 24GB/12GB, and training time of 1.0h/0.5h, as shown in Table \ref{tab:effi}. The top row of each group represents the default setting of the corresponding baseline, where the training time is measured by us with the same number of iterations on a single GPU. While both FreeNeRF and SparseNeRF perform worse under strict resource limitations, our method shows huge advantages in efficiency, which achieves remarkable accelerations of $25\times$ on training time and over $3000\times$ on FPS, while synthesizing competitive quality novel views. Given the necessity for per-scene optimization and rapid visualization, our high efficiency holds significant value for practical applications.

\begin{table}[]
\setlength{\abovecaptionskip}{5pt}
\resizebox{1\linewidth}{!}{
\begin{tabular}{l|c|c|c|ccc}
\toprule
Method                      & FPS                & Time                   & GPU Mem       & PSNR$\uparrow$  & LPIPS$\downarrow$          & SSIM$\uparrow$  \\ \midrule
\multirow{3}{*}{FreeNeRF \cite{yang2023freenerf}}   & \multirow{3}{*}{9$\times$10$^{-2}$} & 2.3 h                  & 4$\times$48 GB        & 19.63 & 0.308          & 0.612 \\
                            &                    & 2.3 h                  & 24 GB         & 19.71 & 0.322          & 0.603 \\
                            &                    & 1.0 h                  & 24 GB         & 19.66 & 0.357          & 0.574 \\ \midrule
\multirow{3}{*}{SparseNeRF \cite{wang2023sparsenerf}} & \multirow{3}{*}{9$\times$10$^{-2}$} & 1.5 h          & 32 GB         & 19.86 & 0.328          & \textbf{0.624} \\
                            &                    & 1.5 h                   & 12 GB         & \textbf{19.95} & 0.334          & 0.598 \\
                            &                    & 0.5 h                  & 12 GB         & 19.94 & 0.341          & 0.585 \\ \midrule
Ours                        & \textbf{300}  & \textbf{3.5 min}       & \textbf{2 GB} & 19.12 & \textbf{0.294} & 0.591 \\ \bottomrule
\end{tabular}
}
\caption{\textbf{Efficiency Comparison with Limited Resources.} Our method achieves efficient training and the fastest real-time rendering while synthesizing competitive high-quality novel views. }
\label{tab:effi}
\vspace{-.2cm}
\end{table}

\begin{table}[]
\setlength{\abovecaptionskip}{5pt}
\setlength{\tabcolsep}{3.5 mm}
\resizebox{1\linewidth}{!}{
\begin{tabular}{cccc|ccc}
\toprule
\multicolumn{2}{c}{Regularization}  & \multicolumn{2}{c|}{Normalization} & \multirow{2}{*}{PSNR $\uparrow$} & \multirow{2}{*}{LPIPS $\downarrow$} & \multirow{2}{*}{SSIM $\uparrow$} \\
AP & \multicolumn{1}{c|}{Hard Soft} & Local           & Global           &                       &                        &                       \\ \midrule
\checkmark  & \multicolumn{1}{c|}{}          &                 &                  & 18.14                 & 0.354                  & 0.538                 \\ 
   & \multicolumn{1}{c|}{\checkmark}         &                 &                           & 17.90                                         & 0.351                                          & 0.525                                         \\ \midrule
\checkmark  & \multicolumn{1}{c|}{}          &                 & \checkmark                & 18.31                 & 0.339                  & 0.552                 \\
\checkmark  & \multicolumn{1}{c|}{}          & \checkmark               & \checkmark                & \cellcolor[HTML]{FFE4CF}18.68                 & \cellcolor[HTML]{FFFFD4}0.331                  & \cellcolor[HTML]{FFE4CF}0.565                 \\ \midrule
   & \multicolumn{1}{c|}{\checkmark}         &                 & \checkmark                & \cellcolor[HTML]{FFFFD4}18.55                 & \cellcolor[HTML]{FFE4CF}0.324                  & \cellcolor[HTML]{FFFFD4}0.562                 \\
   & \multicolumn{1}{c|}{\checkmark}         & \checkmark               & \checkmark                & \cellcolor[HTML]{FFCCC9}19.12                 & \cellcolor[HTML]{FFCCC9}0.294                  & \cellcolor[HTML]{FFCCC9}0.591                 \\ \bottomrule
\end{tabular}
}
\caption{\textbf{Ablation Study.} We ablate our method on the LLFF 3-view setting. The results show the effect of our contributions.}
\label{tab:ablation}
\vspace{-.2cm}
\end{table}

\begin{figure}[t]
\setlength{\abovecaptionskip}{3pt}
    \centering
    \includegraphics[width=1\linewidth]{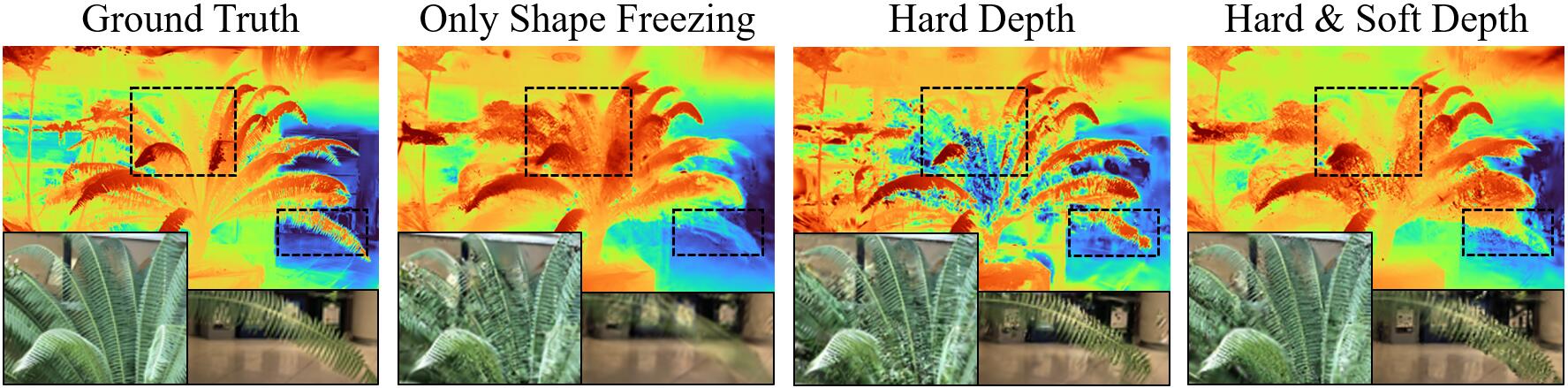}
    \caption{\textbf{Visualization Results of Depth Regularization.} Our Hard Depth Regularization significantly improves the high-frequency details but causes hollows. This drawback can be solved via the Soft Depth to synthesize fine details. We take the depth from dense views as the ground truth for comparison.}
    \label{fig:regular}
    \vspace{-.5cm}
\end{figure}

\vspace{-0.1cm}
\subsection{Ablation Study}
\vspace{-0.1cm}
We ablate our method on the LLFF 3-view setting. The quantitative results are reported in Table \ref{tab:ablation} and \ref{tab:freeze}. 

\vspace{2pt}\noindent\textbf{Depth Regularization.}
We ablate our Hard and Soft Depth Regularization with a plain all-parameter (AP) L2 reconstruction loss term. To better separately illustrate the effect of our two types of depth and exclude the influence of shape freezing, we further visualize a comparison to the situation only with shape freezing in Figure \ref{fig:regular}. It has been shown that the plain depth regularization can not effectively reshape the scene geometry, which proves the necessity of our method. Both the qualitative and quantitative results demonstrate our effect on geometry quality and high-frequency details.

\vspace{2pt}\noindent\textbf{Global-Local Depth Normalization.}
From the result, we can observe that only adding a global normalization can also help fitting, which is mainly due to the local patch-wise loss computation. After the attendance of local normalization, the rendering quality improves especially in detail. These improvements are much more obvious when applied to our proposed regularization than the all-parameter regularization that is unsuitable for the fields. The results correspond to our design and show the effectiveness of our Global-Local Depth Normalization.

\begin{figure}[t]
    \centering
    \setlength{\abovecaptionskip}{5pt}
    \includegraphics[width=1\linewidth]{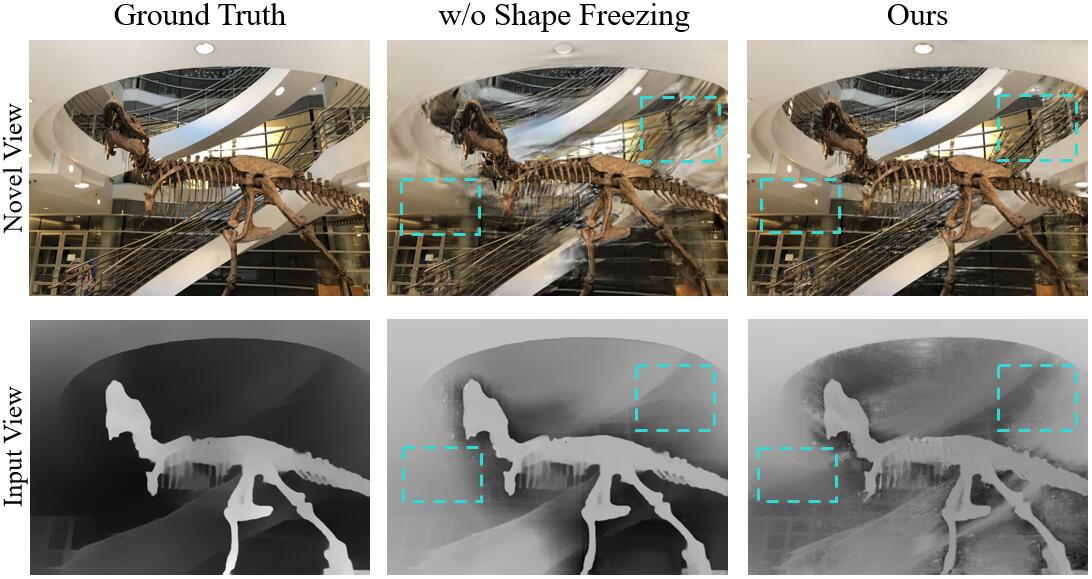}
    \caption{\textbf{Visualization Results of Shape Freezing.} The synthesized novel view without shape freezing is filled with blurry areas, which is mainly caused by the over-smooth geometry learned from the monocular depth (left bottom).}
    \label{fig:freeze}
    \vspace{-.1cm}
\end{figure}

\vspace{2pt}\noindent\textbf{Parameter Freezing.}
To illustrate the effect of our parameter-freezing strategy, we also ablate the shape freezing in regularization and center freezing in soft depth calculation. The results are shown in Table \ref{tab:freeze} and Figure \ref{fig:freeze}. The visualization illustrates the problem of the depth-color conflict in Sec.\ref{sec:reg}. In the situation without center freezing, some primitives may move to unexpected places to compensate for the depth loss, which causes lower quality. By introducing the proposed parameter freezing, we successfully relieve the problems and achieve the best results.

\begin{table}[t]
\setlength{\abovecaptionskip}{5pt}
\resizebox{1\linewidth}{!}{
\setlength{\tabcolsep}{3.2 mm}

\begin{tabular}{l|cccc}
\toprule
Setting             & PSNR $\uparrow$  & LPIPS $\downarrow$ & SSIM $\uparrow$ & AVGE $\downarrow$  \\ \midrule
w/o Shape Freezing  & 17.96 & 0.363 & 0.547 & 0.160 \\
w/o Center Freezing & 18.87 & 0.300 & 0.584 & 0.140 \\
All                & \textbf{19.12} & \textbf{0.294} & \textbf{0.591} & \textbf{0.132} \\ \bottomrule
\end{tabular}
}
\caption{\textbf{Ablation Study on Parameter Freezing.} The results demonstrate the necessity of our parameter freezing strategy.}
\label{tab:freeze}
\vspace{-.5cm}
\end{table}

\vspace{-0.1cm}
\section{Conclusion}
\vspace{-0.1cm}
This paper presents the DNGaussian framework that introduces 3DGS into the few-shot novel view synthesis task by depth regularization. Due to the space limitation, we have put more discussions in the supplementary material.

\vspace{2pt}\noindent\textbf{Acknowledgements.}
In this work, we are supported by the National Natural Science Foundation of China 62276016, 62372029. Lin Gu is supported by JST Moonshot R\&D Grant Number JPMJMS2011 Japan.

%

\clearpage
\setcounter{page}{1}
\maketitlesupplementary

\appendix

\section*{Overview}
In the supplemental document, we first report additional studies in Sec. \ref{sec:results} of our proposed depth normalization, neural color renderer, and the performance of previous methods on fast grid-based backbones. Then, we describe the details of our implementation and dataset settings in our experiment in Sec. \ref{sec:details}. Finally, we discuss the limitations and future work of our method in Sec. \ref{sec:limit}.

\section{Additional Results \label{sec:results}}

\subsection{Ablation Study on Depth Normalization}
To better illustrate the roles of our Local and Global Depth Normalization, we conduct an additional ablation study and replace the L2 loss function with L1 to avoid its reduction of small losses. The quantitative visualization results are shown in Table \ref{tab:norm} and Figure \ref{fig:norm}. In the comparison, we separately apply the Global and the Local one to illustrate the strengths and weaknesses of each: 1) Although the global one can also individually support the model to learn an overall scene, it is weak in optimizing minor errors, as we have discussed in Sec.3.3. 2) The local one can not stand alone due to the lack of absolute scale, but provides rich information on local depth changes. 3) By combining both techniques, our Global-Local Depth Normalization can simultaneously obtain the knowledge of both global scale and small local errors and achieve the best. Notably, since a different type of loss is used in this study, the scores vary from those reported in the main paper. Despite this, our method still performs well particularly in LPIPS and SSIM, which demonstrates the robustness of our depth normalization.

\begin{table}[h]
\setlength{\abovecaptionskip}{4pt}
\resizebox{1\linewidth}{!}{
\setlength{\tabcolsep}{3.5 mm}

\begin{tabular}{l|cccc}
\toprule
Setting             & PSNR$\uparrow$  & LPIPS$\downarrow$ & SSIM$\uparrow$ & AVGE$\downarrow$  \\ \midrule
Only Global  & 18.32 & 0.309 & 0.579 & 0.144 \\
Only Lobal & 17.17 & 0.338 & 0.523 & 0.167 \\
All                & \textbf{18.67} & \textbf{0.291} & \textbf{0.595} & \textbf{0.137} \\ \bottomrule
\end{tabular}
}
\caption{\textbf{Additional Ablation Study on Depth Normalization.} Combined with both two proposed depth normalizations, our Global-Local Depth Normalization achieves the best quality.}
\label{tab:norm}
\vspace{-.4cm}
\end{table}

\subsection{Ablation Study on Neural Color Renderer}
\begin{figure}[t]
    \setlength{\abovecaptionskip}{4pt}
    \centering
    \includegraphics[width=1\linewidth]{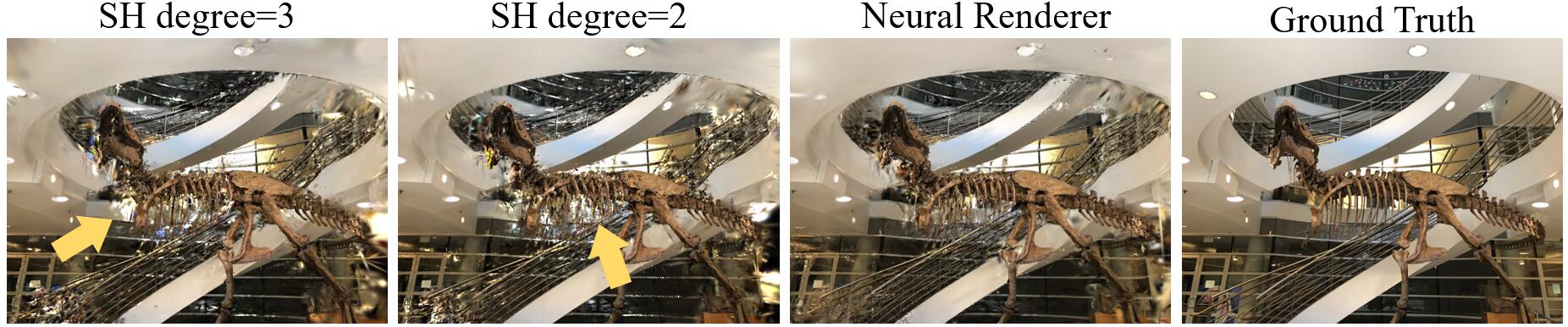}
    \caption{\textbf{Visualization of two proposed Depth Normalization.} In sparse-view situations, SH may produce inconsistent colors in unseen views (yellow arrow) due to overfitting, which can be relieved by a neural color renderer.}
    \label{fig:sh}
\end{figure}

In this work, we replace the spherical harmonic (SH) of 3D Gaussian Splatting with a neural color renderer to represent the direction-variant color. To better illustrate the function of this module, we compare it to the original SH function with different degrees in the LLFF dataset with 3 training views. The results are in Table \ref{tab:sh} and Figure \ref{fig:sh}. The SH function is easy to overfit in the sparse-view situation and results in some strange colors during view changing. This may be caused by the independence of each primitive which leads to a lack of regional consistency. After introducing the neural color renderer, the problem has been relieved. By storing the intermediate result and only calculating the latest two MLP layers, we can maintain a fast rendering speed competitive to SH as well.

\begin{figure}[t]
    \setlength{\abovecaptionskip}{3pt} 
    \centering
    \includegraphics[width=1\linewidth]{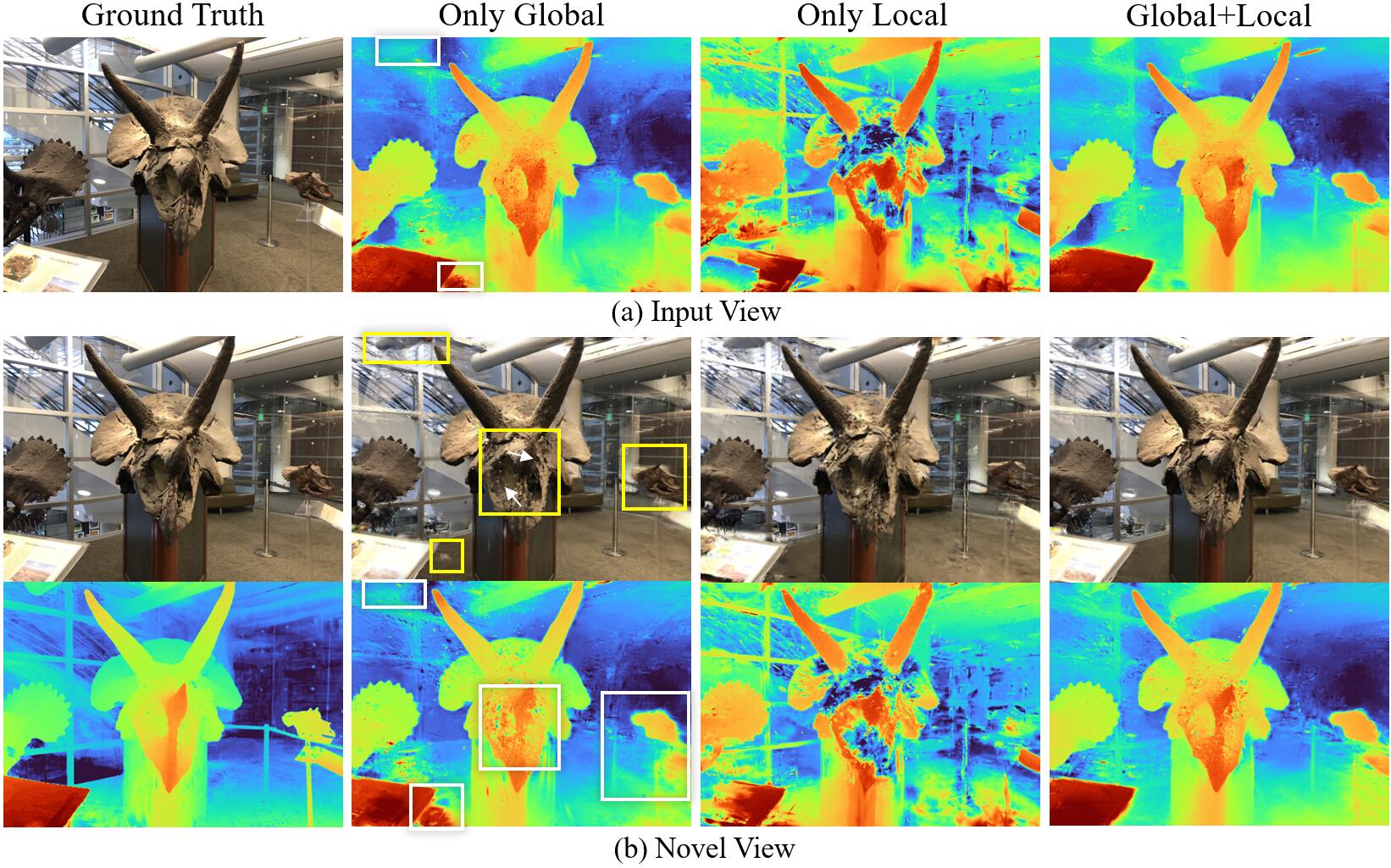}
    \caption{\textbf{Visualization of two proposed Depth Normalization.} The color and depth map of the input view and synthesized novel view are shown in (a) and (b). The global one provides a global view of the whole scene, however, is weak in handling small local errors (white box), which causes blurry and wrong appearances (yellow box). In contrast, the local one is more sensitive to local depth changes. By combining both of them, our method can learn a more accurate scene geometry. Zoom in for better visualization.}
    \label{fig:norm}
\end{figure}

\begin{table}[t]
\setlength{\abovecaptionskip}{4pt}
\resizebox{1\linewidth}{!}{
\setlength{\tabcolsep}{3.2 mm}

\begin{tabular}{l|cccc|c}
\toprule
Setting             & PSNR$\uparrow$  & LPIPS$\downarrow$ & SSIM$\uparrow$ & AVGE$\downarrow$ & FPS  \\ \midrule
SH degree=2  & 17.06 & 0.333 & 0.549 & 0.167 & \textbf{340} \\
SH degree=3  & 17.11 & 0.328 & 0.560 & 0.164 & 300\\ \midrule
Neural Renderer & \textbf{19.12} & \textbf{0.294} & \textbf{0.591} & \textbf{0.132} & 300\\ \bottomrule
\end{tabular}
}
\caption{\textbf{Ablation Study on Neural Color Renderer.} Our neural color renderer successfully improves the rendering quality while keeping an equally fast inference speed.}
\label{tab:sh}
\vspace{-.2cm}
\end{table}

\begin{table*}[t]
\setlength{\abovecaptionskip}{5pt}
\resizebox{1\linewidth}{!}{
\setlength{\tabcolsep}{6.2 mm}
\begin{tabular}{l|l|ccc|ccc}
\toprule
Backbone                     & Strategy   & PSNR $\uparrow$  & LPIPS $\downarrow$ & SSIM $\uparrow$  & Time $\downarrow$ & VM Cost $\downarrow$          & FPS  $\uparrow$           \\ \midrule
\multirow{3}{*}{Mip-NeRF \cite{barron2021mip}}    
                             & None       & 14.62 & 0.495 & 0.351 & 2.2h   & \multirow{3}{*}{$\geq$ 32 GB} & \multirow{3}{*}{0.09} \\
                             & FreeNeRF   & 19.63 & \underline{0.308} & 0.612 & 2.3h   &                         &                       \\
                             & SparseNeRF & \underline{\textbf{19.86}} & 0.328 & \underline{\textbf{0.624}} & 1.5h   &                         &                       \\ \midrule
\multirow{3}{*}{Instant-NGP \cite{muller2022instant}} 
                             & None       & 17.19 & 0.483 & 0.469 & 3.8min   & \multirow{3}{*}{3 GB}   & \multirow{3}{*}{3} \\
                             & FreeNeRF   & 15.30 & 0.516 & 0.369 & 4.2min   &                         &                       \\
                             & SparseNeRF & \underline{17.19} & \underline{0.478} & \underline{0.476} & 7.5min   &                         &                       \\ \midrule
\multirow{3}{*}{TensoRF \cite{chen2022tensorf}}    
                             & None       & \underline{16.16} & \underline{0.454} & \underline{0.443} & 4.1min   & \multirow{3}{*}{8 GB}   & \multirow{3}{*}{5} \\
                             & FreeNeRF   & 15.78 & 0.466 & 0.430 & 4.5min   &                         &                       \\
                             & SparseNeRF & 16.11 & 0.465 & 0.443 & 8.9min   &                         &                       \\ \midrule
\multirow{3}{*}{3DGS \cite{kerbl20233d}}       
                             & None       & 16.46 & 0.401 & 0.440 & \textbf{2.7min}   & \multirow{3}{*}{2 GB}   & \multirow{3}{*}{280}  \\
                             & FreeNeRF   & 16.55 & 0.399 & 0.472 & 2.7min   &                         &                       \\
                             & SparseNeRF & 16.80 & 0.374 & 0.504 & 2.9min   &                         &                       \\ \midrule
3DGS \cite{kerbl20233d}      & Ours       & \underline{19.12} & \underline{\textbf{0.294}} & \underline{0.591} & 3.5min & \textbf{2 GB}                    & \textbf{300}                   \\ \bottomrule
\end{tabular}
}
\caption{\textbf{Comparision of SOTA strategies FreeNeRF \cite{yang2023freenerf} and SparseNeRF \cite{wang2023sparsenerf} with different backbones}. The best results for all and for each backbone are marked with \textbf{bold} and \underline{underline}. Although FreeNeRF and SparseNeRF perform well on the implicit Mip-NeRF \cite{barron2021mip}, they can weakly improve the quality with current fast backbones Instant-NGP \cite{muller2022instant}, TensoRF \cite{chen2022tensorf}, and also the 3DGS \cite{kerbl20233d} in our work.}
\label{tab:backbone}
\end{table*}

\begin{figure*}
    \setlength{\abovecaptionskip}{6pt}
    \centering
    \includegraphics[width=1\linewidth]{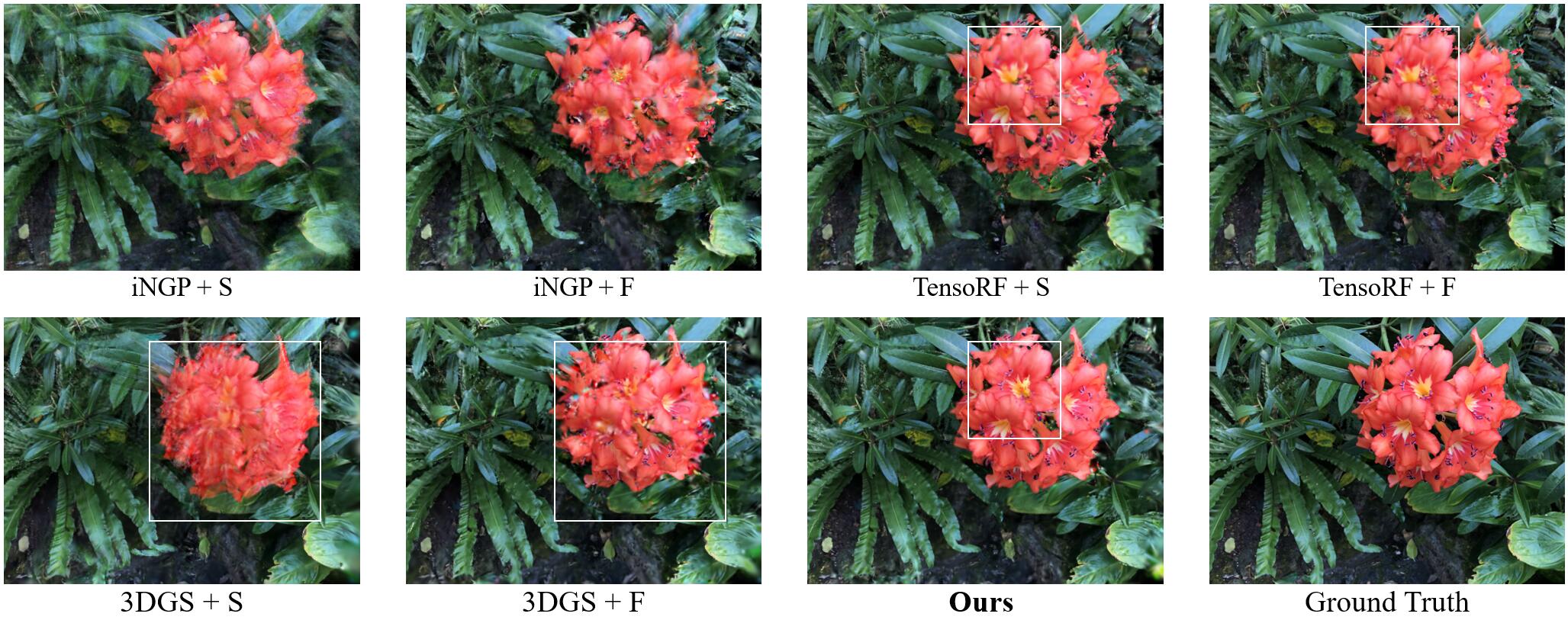}
    \caption{Visualization of the implicit-based SOTA methods SparseNeRF (S) \cite{wang2023sparsenerf} and FreeNeRF (F) \cite{yang2023freenerf} when transferred to fast backbones Instant-NGP (iNGP) \cite{muller2022instant}, TensoRF \cite{chen2022tensorf}, and 3D Gaussian Splatting (3DGS) \cite{kerbl20233d}. The depth-based SparseNeRF causes blurry rendering on all fast backbones, while FreeNeRF has less effect on the improvement of quality. Instead, our DNGaussian achieves the best quality on the most efficient 3DGS backbone, which shows the necessity of our method.}
    \label{fig:backbone}
    \vspace{-.4cm}
\end{figure*}

\subsection{Transfer of Previous Strategies}
In this section, we conduct an experiment to illustrate the necessity of our efficient DNGaussian. Indeed, there are some existing methods like FreeNeRF \cite{yang2023freenerf} and SparseNeRF \cite{wang2023sparsenerf} that are low in efficiency mainly due to their backbone rather than the strategy itself. However, they have only already been proven effective for some implicit backbones that are slow and costly. To verify whether they can directly transfer to faster backbones to achieve higher efficiency, we implement these two methods on two fast grid-based Instant-NGP \cite{muller2022instant} and TensoRF \cite{chen2022tensorf}. Also, we do this on our 3D Gaussian Splatting (3DGS) \cite{kerbl20233d} backbone. Then, we test these implementations in the LLFF 3-view setting. The results are shown in Table \ref{tab:backbone} and Figure \ref{fig:backbone}. Additionally, we report the training time (Time), GPU memory cost (VM Cost), and the inference FPS for each item.

\vspace{2pt}\noindent\textbf{Implementation details.} We utilize a CUDA-implemented ray marching \footnote{https://github.com/ashawkey/torch-ngp} for the two grid-based backbones to achieve faster speed and lower costs. The 3DGS backbone employs the same neural color renderer as our method. We follow the original implementation of SparseNeRF to produce monocular depth maps for all input views and transfer its Local Depth Ranking Distillation to these new backbones with the same hyperparameters. For FreeNeRF, since the three fast backbones do not contain a frequency-based positional encoding, we apply the Frequency Regularization to their grid-based positional encoding as an alternative.

\vspace{2pt}\noindent\textbf{Comparison on grid-based backbones.} Although these two methods perform well on their original implicit Mip-NeRF, they are weak in both Instant-NGP and TensoRF. SparseNeRF distills the depth ranking from the monocular depth map for regularization, however, causes more blurs. This may be caused by the stronger spatial memory ability from the explicit grids that makes it easier to memorize noises. FreeNeRF performs even worse on both these two backbones, which may be due to the different representations of positional encoding. In TensoRF, all these two strategies fail to improve performance. One reason may lie in that TensoRF directly utilizes explicit grids without a neural decoder to store density value, which is more difficult to regularize.

\vspace{2pt}\noindent\textbf{Comparison on 3DGS.} In the comparison, the 3DGS backbone achieves the best efficiency, with the fast FPS and lowest training cost. However, both SparseNeRF and FreeNeRF cannot effectively regularize this powerful and efficient backbone. Due to the lack of frequency positional encoding, FreeNeRF serves more like a coarse-to-fine strategy and leads to only a little improvement. From the visualization of SparseNeRF in Figure \ref{fig:backbone}, it can be observed that it is insufficient in the 3D Gaussian radiance fields of 3DGS to only keep the depth ranking and wait for the color-supervised optimization process to refine the detailed geometry. Compared with these two methods, our DNGaussian achieves a much better quality with only a little increment of training time. With less noise in the learned geometry, our method also achieves a faster inference speed.

\vspace{2pt}\noindent\textbf{Conclusion.} The experiments show that the previous methods for implicit backbones can hardly, at least in an easy way, transfer to current fast backbones. Also, they are not suitable for the 3D Gaussian radiance fields. In such a situation, our DNGaussian shows significant value in providing an efficient way for high-quality and low-cost few-shot novel view synthesis.

\subsection{Comparison with Grid-based Methods}

There are some works \cite{wynn2023diffusionerf, sun2023vgos, song2023darf} that utilize a grid-based backbone to improve the training efficiency. Since DaRF \cite{song2023darf} is evaluated on another two datasets with at least 9 input views, while VGOS \cite{sun2023vgos} and DiffusioNeRF \cite{wynn2023diffusionerf} use different methods for the measurement of metrics, we do not take them as baselines in the main paper. Here we list the scores of VGOS and DiffusioNeRF in Table \ref{tab:grid} in the LLFF 3-view setting for comparison. For VGOS, we only report scores for which the measurement method is definitely the same as in RegNeRF \cite{niemeyer2022regnerf} and FreeNeRF \cite{yang2023freenerf}. The results of DiffusioNeRF are obtained from its updated paper on arXiv \footnote{https://arxiv.org/abs/2302.12231}. In the comparison, our method outperforms the other two with the highest scores in LPIPS, SSIM, and AVGE. In fact, our method also achieves the best in efficiency, with much lower cost and faster inference.

\subsection{Additional Visualizations}
\begin{figure}[t]
    \setlength{\abovecaptionskip}{6pt}
    \centering
    \includegraphics[width=1\linewidth]{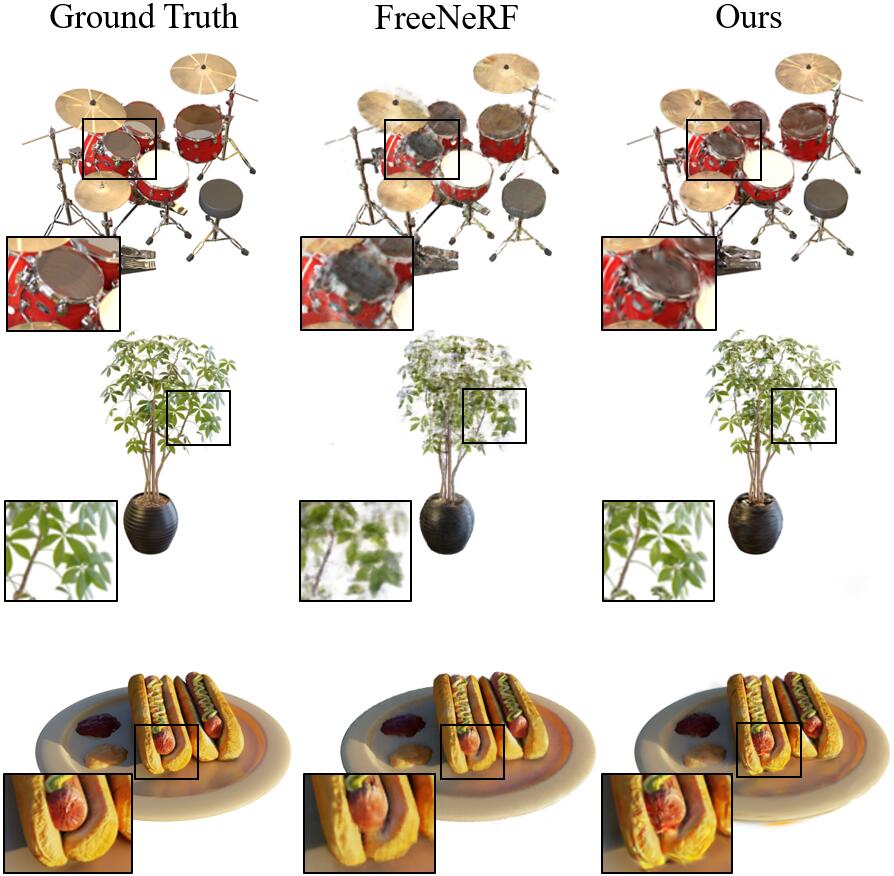}
    \caption{\textbf{Qualitative comparison on the Blender dataset with 8 input views.} FreeNeRF \cite{yang2023freenerf} learns the accurate geometry by masking high-frequency signals, however, suffers from blurry details as the trade-off. In contrast, our method does not explicitly constrain the learning of high-frequency content. Also attributed to the 3D Gaussian neural fields, our method performs better in the fine-grained details.}
    \label{fig:blender_more}
\end{figure}

We provide more rendering results in our experiments. The examples on DTU and LLFF with 3 training views are shown in Figure \ref{fig:dtu_all} and \ref{fig:llff_all}. We have also shown more quantitative comparison in the Blender 8-view setting with the SOTA method FreeNeRF \cite{yang2023freenerf} in Figure \ref{fig:blender_more}. More results can be found in our \textcolor{blue}{supplementary video}.

\begin{table}[t]
\setlength{\abovecaptionskip}{4pt}
\resizebox{1\linewidth}{!}{
\setlength{\tabcolsep}{3.5 mm}

\begin{tabular}{l|cccc}
\toprule
Method             & PSNR$\uparrow$  & LPIPS$\downarrow$ & SSIM$\uparrow$ & AVGE$\downarrow$  \\ \midrule
VGOS \cite{sun2023vgos} & 19.35 & 0.432 & - & - \\
DiffusioNeRF \cite{wynn2023diffusionerf} & \textbf{19.79} & 0.338 & 0.568 & 0.136 \\ \midrule
Ours  & 19.12 & \textbf{0.294} & \textbf{0.591} & \textbf{0.132} \\ \bottomrule
\end{tabular}
}
\caption{\textbf{Comparison with grid-based few-shot NeRFs on LLFF with 3 training views.} Our method outperforms grid-based methods VGOS \cite{sun2023vgos} and DiffusioNeRF \cite{sun2023vgos}.}
\label{tab:grid}
\vspace{-.4cm}
\end{table}

\begin{figure*}[!t]
    \vspace{1cm}
    \setlength{\abovecaptionskip}{12pt}
    \centering
    \includegraphics[width=1\linewidth]{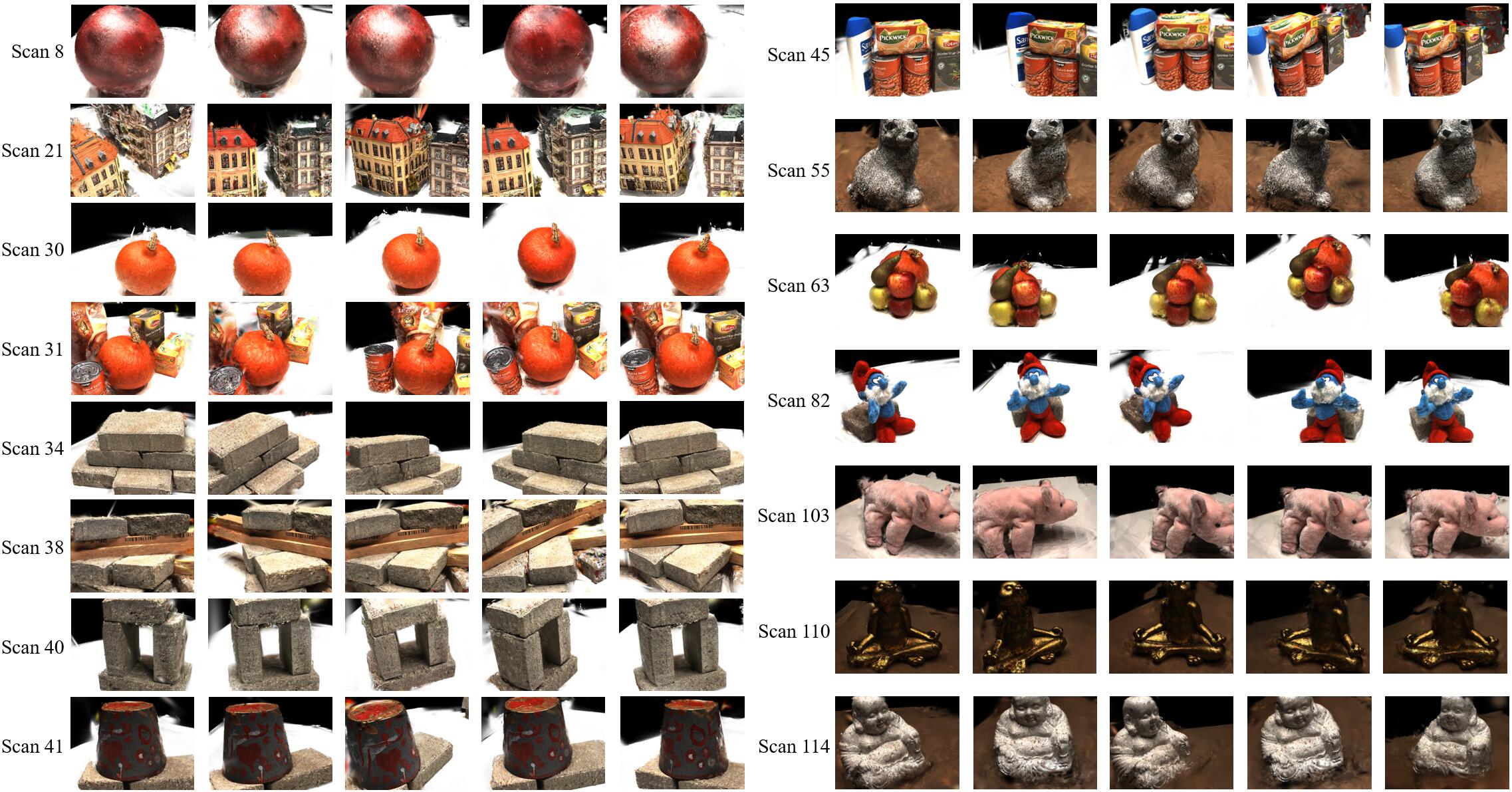}
    \caption{\textbf{Examples of the synthesized novel view results from DNGaussian with 3 input views on the DTU dataset.}}
    \label{fig:dtu_all}
    \vspace{1cm}
\end{figure*}

\begin{figure*}[!t]
    \setlength{\abovecaptionskip}{12pt}
    \centering
    \includegraphics[width=1\linewidth]{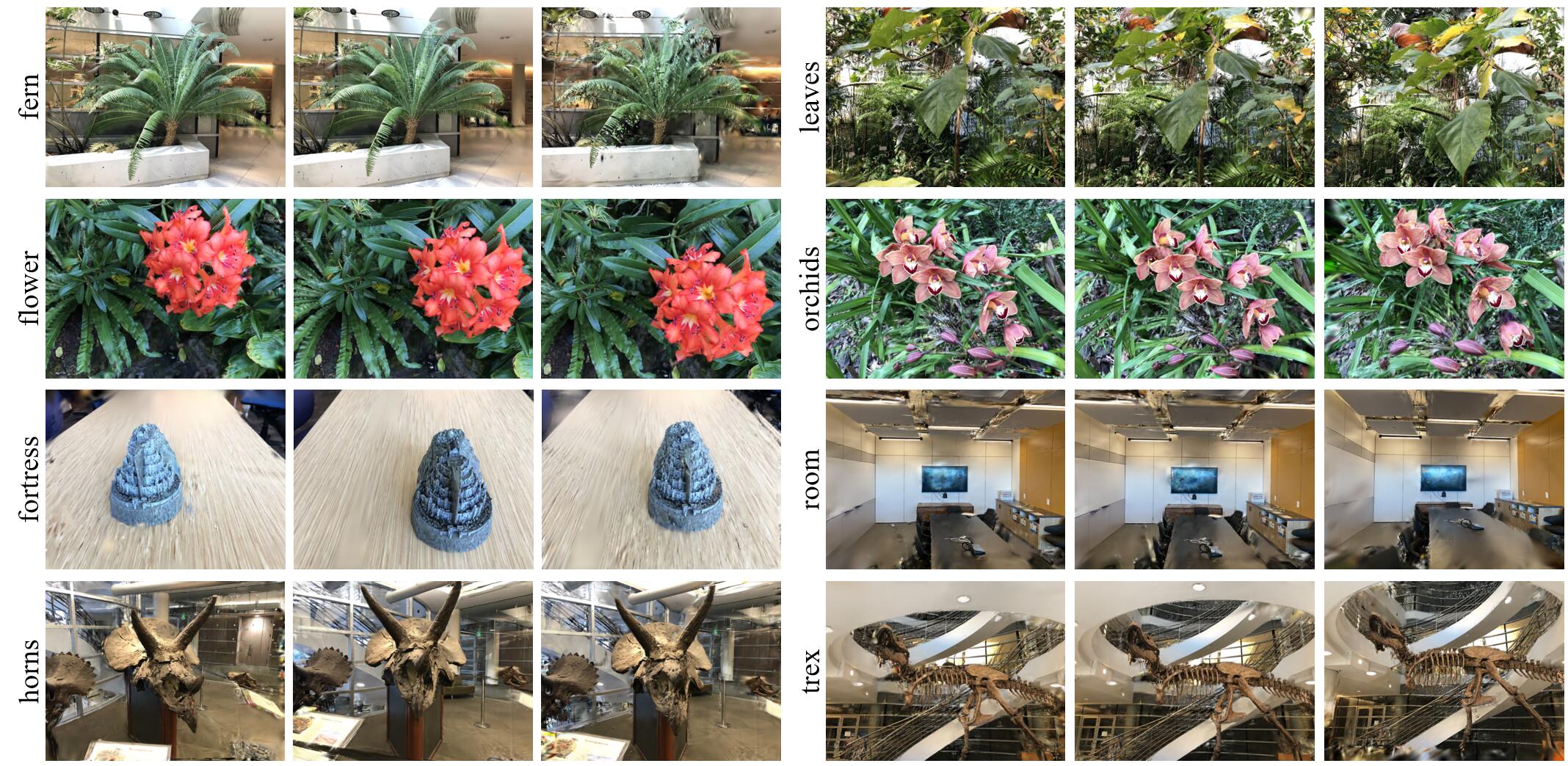}
    \caption{\textbf{Examples of the synthesized novel view results from DNGaussian with 3 input views on the LLFF dataset.}}
    \label{fig:llff_all}
\end{figure*}

\section{Details \label{sec:details}}
\subsection{Implementations}
\vspace{2pt}\noindent\textbf{Pre-trained Depth Models.} 
In this work, we use the pre-trained DPT \cite{Ranftl2021dpt, Ranftl2022midas} estimator to predict the depth map, which has been widely used in many NeRF-based works \cite{deng2023nerdi, wang2023sparsenerf, bian2023nope, wu2022dof}. Particularly, we use the type of \textit{dpt\_hybrid\_384} for the LLFF dataset, while \textit{dpt\_large\_384} for DTU and Blender, which performs better for the pure white or black background. In fact, the performance gaps of our method when applying different types of depth models are slight, as shown in Table \ref{tab:depth}.

\begin{table}[t]
\setlength{\abovecaptionskip}{4pt}
\resizebox{1\linewidth}{!}{
\setlength{\tabcolsep}{4 mm}
\begin{tabular}{l|cccc}
\toprule
Type             & PSNR $\uparrow$  & LPIPS $\downarrow$ & SSIM $\uparrow$ & AVGE $\downarrow$  \\ \midrule
 \multicolumn{5}{c}{\textbf{\quad \quad \quad LLFF}} \\  \midrule
dpt\_hybrid\_384* & 19.12 & 0.294 & 0.591 & 0.132 \\
dpt\_large\_384  & 19.03 & 0.297 & 0.590 & 0.135 \\ \midrule
 \multicolumn{5}{c}{\textbf{\quad \quad \quad DTU}} \\  \midrule
dpt\_hybrid\_384 & 18.86 & 0.179 & 0.784 & 0.106 \\
dpt\_large\_384*  & 18.91 & 0.176 & 0.790 & 0.102 \\ \bottomrule
\end{tabular}
}
\caption{\textbf{The influence of different pre-trained depth models.} We replace the pre-trained depth model with a different type in our  LLFF and DTU settings while keeping the same hyperparameters. The results show that our method is robust to different monocular depth estimators. * denotes the default type of each dataset.}
\label{tab:depth}
\vspace{-.2cm}
\end{table}

\vspace{2pt}\noindent\textbf{Patch Size.} 
In our implementation, we randomly sample a patch size from a pre-defined range for our patch-based Global-Local Depth Normalization. This range is set of $[5, 17]$ for LLFF and Blender, and a larger $[17, 51]$ for DTU since the objects are smaller but occupy a large proportion of the image. Due to the flexibility of our normalization, we do not need to separately tune this value for each scene.


\vspace{2pt}\noindent\textbf{Metrics.} 
Following previous methods \cite{niemeyer2022regnerf, yang2023freenerf}, we utilize the ``structural\_similarity" API from scikit-image  \footnote{https://scikit-image.org/docs/stable/api/skimage.metrics.html} to compute the SSIM score, and use the implementation with a pre-trained VGG model to calculate the LPIPS score.

\vspace{2pt}\noindent\textbf{Camera Poses.}
Following existing works \cite{niemeyer2022regnerf, yang2023freenerf, wang2023sparsenerf, jain2021putting}, we assume all camera poses are already known. In practice, for LLFF and Blender, we use the given poses from the datasets. For DTU, we use COLMAP \cite{schoenberger2016mvs, schoenberger2016sfm} to calculate the camera poses according to all given views, and then sample the target sparse views from the results. 

\subsection{Datasets}
\vspace{2pt}\noindent\textbf{LLFF.}
The LLFF dataset \cite{mildenhall2019local} contains 8 forward-facing scenes in total. Following \cite{niemeyer2022regnerf, yang2023freenerf, wang2023sparsenerf}, we take every 8-th image as the novel views for testing. The input views are evenly sampled across the remaining views. Images are downsampled $8\times$ to the resolution of  $378\times504$. In practice, we ignore the distortion of the original images.

\vspace{2pt}\noindent\textbf{DTU.}
The DTU dataset \cite{jensen2014large} consists of 124 object-centric scenes captured by a set of fixed cameras. We follow \cite{niemeyer2022regnerf, yang2023freenerf, wang2023sparsenerf} to evaluate models directly on the 15 scenes with the scan IDs of 8, 21, 30, 31, 34, 38, 40, 41, 45, 55, 63, 82, 103, 110, and 114. In each scan, the images with the following IDs of 25, 22, and 28 are used as the input views in our 3-view setting. The test set consists of images with IDs of 1, 2, 9, 10, 11, 12, 14, 15, 23, 24, 26, 27, 29, 30, 31, 32, 33, 34, 35, 41, 42, 43, 45, 46 and 47 for evaluation. The images are downsampled $4\times$. In particular, we use the undistorted images from COLMAP to eliminate the negative impact of unerased lens distortion.

\vspace{2pt}\noindent\textbf{Blender.}
We follow the data split used in \cite{jain2021putting, yang2023freenerf} for the Blender dataset \cite{deng2022dsnerf}. The 8 input views are selected from the training images, with IDs 26, 86, 2, 55, 75, 93, 16, 73, and 8. The 25 test views are sampled evenly from the testing images for evaluation. All images are downsampled $2\times$ to $400 \times 400$ during the experiment.

\section{Discussions and Limitations \label{sec:limit}}
Our DNGaussian utilizes coarse monocular depth to regularize the scene geometry in situations with sparse input views, and achieves significant improvement in the appearance quality. However, our method still has limitations such as below. We hope these issues can be solved in future work.

\vspace{2pt}\noindent\textbf{More Input Views.}
 Besides only 3 input views, we have also explored the performance when the number of input views increases to 6 and 9 on the LLFF dataset, as shown in Table \ref{tab:moreview}. In the experiment, it can be observed that as the number of views increases, the performance of the baseline also improves. Our DNGaussian can still improve the quality of the synthesized novel view with 6 input views. However, it does not work well when the number of input views increases to 9, which is nearly enough to provide sufficient color constraints. This may be due to the errors in the depth map that negatively influence the optimization process. The next step of our work can lie in leveraging the uncertainty of the monocular depth to filter out unreliable supervision.

\begin{table}[h]
\setlength{\abovecaptionskip}{4pt}
\resizebox{1\linewidth}{!}{
\setlength{\tabcolsep}{3 mm}
\begin{tabular}{c|l|cccc}
\toprule
\multicolumn{1}{l|}{Views} & Method     & PSNR $\uparrow$  & LPIPS $\downarrow$ & SSIM $\uparrow$ & AVGE $\downarrow$  \\ \midrule
\multirow{3}{*}{3}         & 3DGS       & 15.52 & 0.405 & 0.408 & 0.209 \\
                           & 3DGS\dag   & 16.46 & 0.401 & 0.440 & 0.192 \\
                           & DNGaussian & \textbf{19.12} & \textbf{0.294} & \textbf{0.591} & \textbf{0.132} \\ \midrule
\multirow{3}{*}{6}         & 3DGS       & 20.63 & 0.226 & 0.699 & 0.108 \\
                           & 3DGS\dag   & 21.09 & 0.229 & 0.699 & 0.103 \\
                           & DNGaussian & \textbf{22.18} & \textbf{0.198} & \textbf{0.755} & \textbf{0.088} \\ \midrule
\multirow{3}{*}{9}         & 3DGS       & 20.44 & 0.230 & 0.697 & 0.108 \\
                           & 3DGS\dag   & \textbf{23.21} & \textbf{0.176} & 0.785 & \textbf{0.076} \\
                           & DNGaussian & 23.17 & 0.180 & \textbf{0.788} & 0.077 \\ \bottomrule
\end{tabular}
}
\caption{\textbf{Comparison with 3, 6, and 9 input views on LLFF dataset.} \dag \ denotes applied with the same hyperparameters and the neural color renderer as DNGaussian. }
\label{tab:moreview}
\end{table}

\vspace{2pt}\noindent\textbf{Solid Color Planes.}
The anisotropic shape of the Gaussian primitive makes it difficult to represent a solid color plane in a situation with sparse input views. First, the primitives are hard to constrain both by color and depth in the region of the plane, which may cause ray-like noises and hollows. Also, since they can freely move to other regions with similar colors, the densification operation can be activated more frequently and generate noises. This is hoped solved by additional geometry priors.

\vspace{2pt}\noindent\textbf{Specular Regions.}
Although our method can handle some specular regions by relying on depth supervision, especially from our Local Depth Normalization, the inconsistent appearances in these regions are still challenging for 3DGS. To completely solve this problem may still need more special designs.

\vspace{2pt}\noindent\textbf{Hollows and Cracks.}
The splatting technique of our Gaussian Splatting \cite{kerbl20233d} backbone directly merges existing primitives to render the pixel-level color without interpolation. However, since not every pixel can be overlapped by the projected primitives, the empty space between two Gaussian primitives would cause hollows and cracks when the camera pose changes. For example, some hollows can be seen at Scan 40 in Figure \ref{fig:dtu_all}. In this work, we try to solve this problem by paying more attention to high-frequency details and therefore encouraging the densifying of primitives to fill these empty areas. In the future, we believe this problem can be fundamentally solved by the improvement of the representation itself.

{
    \small
    \bibliographystyle{ieeenat_fullname}
    \bibliography{main}

\begin{thebibliography}{63}
\providecommand{\natexlab}[1]{#1}
\providecommand{\url}[1]{\texttt{#1}}
\expandafter\ifx\csname urlstyle\endcsname\relax
  \providecommand{\doi}[1]{doi: #1}\else
  \providecommand{\doi}{doi: \begingroup \urlstyle{rm}\Url}\fi

\bibitem[Avidan and Shashua(1997)]{avidan1997novel}
Shai Avidan and Amnon Shashua.
\newblock Novel view synthesis in tensor space.
\newblock In \emph{Proceedings of IEEE Computer Society Conference on Computer Vision and Pattern Recognition}, pages 1034--1040. IEEE, 1997.

\bibitem[Barron et~al.(2021)Barron, Mildenhall, Tancik, Hedman, Martin-Brualla, and Srinivasan]{barron2021mip}
Jonathan~T Barron, Ben Mildenhall, Matthew Tancik, Peter Hedman, Ricardo Martin-Brualla, and Pratul~P Srinivasan.
\newblock Mip-nerf: A multiscale representation for anti-aliasing neural radiance fields.
\newblock In \emph{Proceedings of the IEEE/CVF International Conference on Computer Vision}, pages 5855--5864, 2021.

\bibitem[Barron et~al.(2022)Barron, Mildenhall, Verbin, Srinivasan, and Hedman]{barron2022mip360}
Jonathan~T Barron, Ben Mildenhall, Dor Verbin, Pratul~P Srinivasan, and Peter Hedman.
\newblock Mip-nerf 360: Unbounded anti-aliased neural radiance fields.
\newblock In \emph{Proceedings of the IEEE/CVF Conference on Computer Vision and Pattern Recognition}, pages 5470--5479, 2022.

\bibitem[Bian et~al.(2023)Bian, Wang, Li, Bian, and Prisacariu]{bian2023nope}
Wenjing Bian, Zirui Wang, Kejie Li, Jia-Wang Bian, and Victor~Adrian Prisacariu.
\newblock Nope-nerf: Optimising neural radiance field with no pose prior.
\newblock In \emph{Proceedings of the IEEE/CVF Conference on Computer Vision and Pattern Recognition}, pages 4160--4169, 2023.

\bibitem[Chen et~al.(2021)Chen, Xu, Zhao, Zhang, Xiang, Yu, and Su]{chen2021mvsnerf}
Anpei Chen, Zexiang Xu, Fuqiang Zhao, Xiaoshuai Zhang, Fanbo Xiang, Jingyi Yu, and Hao Su.
\newblock Mvsnerf: Fast generalizable radiance field reconstruction from multi-view stereo.
\newblock In \emph{Proceedings of the IEEE/CVF International Conference on Computer Vision}, pages 14124--14133, 2021.

\bibitem[Chen et~al.(2022)Chen, Xu, Geiger, Yu, and Su]{chen2022tensorf}
Anpei Chen, Zexiang Xu, Andreas Geiger, Jingyi Yu, and Hao Su.
\newblock Tensorf: Tensorial radiance fields.
\newblock In \emph{Computer Vision--ECCV 2022: 17th European Conference, Tel Aviv, Israel, October 23--27, 2022, Proceedings, Part XXXII}, pages 333--350. Springer, 2022.

\bibitem[Chen et~al.(2023)Chen, Li, Song, Chen, Yu, Yuan, and Xu]{chen2023neurbf}
Zhang Chen, Zhong Li, Liangchen Song, Lele Chen, Jingyi Yu, Junsong Yuan, and Yi Xu.
\newblock Neurbf: A neural fields representation with adaptive radial basis functions.
\newblock In \emph{Proceedings of the IEEE/CVF International Conference on Computer Vision}, pages 4182--4194, 2023.

\bibitem[Chibane et~al.(2021)Chibane, Bansal, Lazova, and Pons-Moll]{chibane2021stereo}
Julian Chibane, Aayush Bansal, Verica Lazova, and Gerard Pons-Moll.
\newblock Stereo radiance fields (srf): Learning view synthesis for sparse views of novel scenes.
\newblock In \emph{Proceedings of the IEEE/CVF Conference on Computer Vision and Pattern Recognition}, pages 7911--7920, 2021.

\bibitem[Cong et~al.(2023)Cong, Liang, Wang, Fan, Chen, Varma, Wang, and Wang]{cong2023enhancing}
Wenyan Cong, Hanxue Liang, Peihao Wang, Zhiwen Fan, Tianlong Chen, Mukund Varma, Yi Wang, and Zhangyang Wang.
\newblock Enhancing nerf akin to enhancing llms: Generalizable nerf transformer with mixture-of-view-experts.
\newblock In \emph{Proceedings of the IEEE/CVF International Conference on Computer Vision}, pages 3193--3204, 2023.

\bibitem[Deng et~al.(2023)Deng, Jiang, Qi, Yan, Zhou, Guibas, Anguelov, et~al.]{deng2023nerdi}
Congyue Deng, Chiyu Jiang, Charles~R Qi, Xinchen Yan, Yin Zhou, Leonidas Guibas, Dragomir Anguelov, et~al.
\newblock Nerdi: Single-view nerf synthesis with language-guided diffusion as general image priors.
\newblock In \emph{Proceedings of the IEEE/CVF Conference on Computer Vision and Pattern Recognition}, pages 20637--20647, 2023.

\bibitem[Deng et~al.(2022)Deng, Liu, Zhu, and Ramanan]{deng2022dsnerf}
Kangle Deng, Andrew Liu, Jun-Yan Zhu, and Deva Ramanan.
\newblock Depth-supervised nerf: Fewer views and faster training for free.
\newblock In \emph{Proceedings of the IEEE/CVF Conference on Computer Vision and Pattern Recognition}, pages 12882--12891, 2022.

\bibitem[Fridovich-Keil et~al.(2022)Fridovich-Keil, Yu, Tancik, Chen, Recht, and Kanazawa]{yu2021plenoxels}
Sara Fridovich-Keil, Alex Yu, Matthew Tancik, Qinhong Chen, Benjamin Recht, and Angjoo Kanazawa.
\newblock Plenoxels: Radiance fields without neural networks.
\newblock In \emph{Proceedings of the IEEE/CVF Conference on Computer Vision and Pattern Recognition}, pages 5501--5510, 2022.

\bibitem[Fridovich-Keil et~al.(2023)Fridovich-Keil, Meanti, Warburg, Recht, and Kanazawa]{fridovich2023k}
Sara Fridovich-Keil, Giacomo Meanti, Frederik~Rahb{\ae}k Warburg, Benjamin Recht, and Angjoo Kanazawa.
\newblock K-planes: Explicit radiance fields in space, time, and appearance.
\newblock In \emph{Proceedings of the IEEE/CVF Conference on Computer Vision and Pattern Recognition}, pages 12479--12488, 2023.

\bibitem[Hu et~al.(2023{\natexlab{a}})Hu, Zhou, Li, Yu, Hong, Hu, Li, Lee, and Liu]{hu2023consistentnerf}
Shoukang Hu, Kaichen Zhou, Kaiyu Li, Longhui Yu, Lanqing Hong, Tianyang Hu, Zhenguo Li, Gim~Hee Lee, and Ziwei Liu.
\newblock Consistentnerf: Enhancing neural radiance fields with 3d consistency for sparse view synthesis.
\newblock \emph{arXiv preprint arXiv:2305.11031}, 2023{\natexlab{a}}.

\bibitem[Hu et~al.(2023{\natexlab{b}})Hu, Wang, Ma, Yang, Gao, Liu, and Ma]{hu2023tri}
Wenbo Hu, Yuling Wang, Lin Ma, Bangbang Yang, Lin Gao, Xiao Liu, and Yuewen Ma.
\newblock Tri-miprf: Tri-mip representation for efficient anti-aliasing neural radiance fields.
\newblock In \emph{Proceedings of the IEEE/CVF International Conference on Computer Vision}, pages 19774--19783, 2023{\natexlab{b}}.

\bibitem[Jain et~al.(2021)Jain, Tancik, and Abbeel]{jain2021putting}
Ajay Jain, Matthew Tancik, and Pieter Abbeel.
\newblock Putting nerf on a diet: Semantically consistent few-shot view synthesis.
\newblock In \emph{Proceedings of the IEEE/CVF International Conference on Computer Vision}, pages 5885--5894, 2021.

\bibitem[Jensen et~al.(2014)Jensen, Dahl, Vogiatzis, Tola, and Aan{\ae}s]{jensen2014large}
Rasmus Jensen, Anders Dahl, George Vogiatzis, Engin Tola, and Henrik Aan{\ae}s.
\newblock Large scale multi-view stereopsis evaluation.
\newblock In \emph{Proceedings of the IEEE conference on computer vision and pattern recognition}, pages 406--413, 2014.

\bibitem[Kerbl et~al.(2023)Kerbl, Kopanas, Leimk{\"u}hler, and Drettakis]{kerbl20233d}
Bernhard Kerbl, Georgios Kopanas, Thomas Leimk{\"u}hler, and George Drettakis.
\newblock 3d gaussian splatting for real-time radiance field rendering.
\newblock \emph{ACM Transactions on Graphics (ToG)}, 42\penalty0 (4):\penalty0 1--14, 2023.

\bibitem[Kim et~al.(2022)Kim, Seo, and Han]{kim2022infonerf}
Mijeong Kim, Seonguk Seo, and Bohyung Han.
\newblock Infonerf: Ray entropy minimization for few-shot neural volume rendering.
\newblock In \emph{Proceedings of the IEEE/CVF Conference on Computer Vision and Pattern Recognition}, pages 12912--12921, 2022.

\bibitem[Kulh{\'a}nek et~al.(2022)Kulh{\'a}nek, Derner, Sattler, and Babu{\v{s}}ka]{kulhanek2022viewformer}
Jon{\'a}{\v{s}} Kulh{\'a}nek, Erik Derner, Torsten Sattler, and Robert Babu{\v{s}}ka.
\newblock Viewformer: Nerf-free neural rendering from few images using transformers.
\newblock In \emph{European Conference on Computer Vision (ECCV)}, 2022.

\bibitem[Liu et~al.(2020)Liu, Gu, Zaw~Lin, Chua, and Theobalt]{liu2020nsvf}
Lingjie Liu, Jiatao Gu, Kyaw Zaw~Lin, Tat-Seng Chua, and Christian Theobalt.
\newblock Neural sparse voxel fields.
\newblock \emph{Advances in Neural Information Processing Systems}, 33:\penalty0 15651--15663, 2020.

\bibitem[Liu et~al.(2023)Liu, Wu, Van~Hoorick, Tokmakov, Zakharov, and Vondrick]{liu2023zero}
Ruoshi Liu, Rundi Wu, Basile Van~Hoorick, Pavel Tokmakov, Sergey Zakharov, and Carl Vondrick.
\newblock Zero-1-to-3: Zero-shot one image to 3d object.
\newblock In \emph{Proceedings of the IEEE/CVF International Conference on Computer Vision}, pages 9298--9309, 2023.

\bibitem[Luiten et~al.(2023)Luiten, Kopanas, Leibe, and Ramanan]{luiten2023dynamic}
Jonathon Luiten, Georgios Kopanas, Bastian Leibe, and Deva Ramanan.
\newblock Dynamic 3d gaussians: Tracking by persistent dynamic view synthesis.
\newblock \emph{arXiv preprint arXiv:2308.09713}, 2023.

\bibitem[Mildenhall et~al.(2019)Mildenhall, Srinivasan, Ortiz-Cayon, Kalantari, Ramamoorthi, Ng, and Kar]{mildenhall2019local}
Ben Mildenhall, Pratul~P Srinivasan, Rodrigo Ortiz-Cayon, Nima~Khademi Kalantari, Ravi Ramamoorthi, Ren Ng, and Abhishek Kar.
\newblock Local light field fusion: Practical view synthesis with prescriptive sampling guidelines.
\newblock \emph{ACM Transactions on Graphics (TOG)}, 38\penalty0 (4):\penalty0 1--14, 2019.

\bibitem[Mildenhall et~al.(2021)Mildenhall, Srinivasan, Tancik, Barron, Ramamoorthi, and Ng]{mildenhall2021nerf}
Ben Mildenhall, Pratul~P Srinivasan, Matthew Tancik, Jonathan~T Barron, Ravi Ramamoorthi, and Ren Ng.
\newblock Nerf: Representing scenes as neural radiance fields for view synthesis.
\newblock \emph{Communications of the ACM}, 65\penalty0 (1):\penalty0 99--106, 2021.

\bibitem[M{\"u}ller et~al.(2022)M{\"u}ller, Evans, Schied, and Keller]{muller2022instant}
Thomas M{\"u}ller, Alex Evans, Christoph Schied, and Alexander Keller.
\newblock Instant neural graphics primitives with a multiresolution hash encoding.
\newblock \emph{ACM Transactions on Graphics (ToG)}, 41\penalty0 (4):\penalty0 1--15, 2022.

\bibitem[Niemeyer et~al.(2022)Niemeyer, Barron, Mildenhall, Sajjadi, Geiger, and Radwan]{niemeyer2022regnerf}
Michael Niemeyer, Jonathan~T Barron, Ben Mildenhall, Mehdi~SM Sajjadi, Andreas Geiger, and Noha Radwan.
\newblock Regnerf: Regularizing neural radiance fields for view synthesis from sparse inputs.
\newblock In \emph{Proceedings of the IEEE/CVF Conference on Computer Vision and Pattern Recognition}, pages 5480--5490, 2022.

\bibitem[Qian et~al.(2023)Qian, Mai, Hamdi, Ren, Siarohin, Li, Lee, Skorokhodov, Wonka, Tulyakov, et~al.]{qian2023magic123}
Guocheng Qian, Jinjie Mai, Abdullah Hamdi, Jian Ren, Aliaksandr Siarohin, Bing Li, Hsin-Ying Lee, Ivan Skorokhodov, Peter Wonka, Sergey Tulyakov, et~al.
\newblock Magic123: One image to high-quality 3d object generation using both 2d and 3d diffusion priors.
\newblock \emph{arXiv preprint arXiv:2306.17843}, 2023.

\bibitem[Ranftl et~al.(2021)Ranftl, Bochkovskiy, and Koltun]{Ranftl2021dpt}
Ren\'{e} Ranftl, Alexey Bochkovskiy, and Vladlen Koltun.
\newblock Vision transformers for dense prediction.
\newblock \emph{ICCV}, 2021.

\bibitem[Ranftl et~al.(2022)Ranftl, Lasinger, Hafner, Schindler, and Koltun]{Ranftl2022midas}
Ren\'{e} Ranftl, Katrin Lasinger, David Hafner, Konrad Schindler, and Vladlen Koltun.
\newblock Towards robust monocular depth estimation: Mixing datasets for zero-shot cross-dataset transfer.
\newblock \emph{IEEE Transactions on Pattern Analysis and Machine Intelligence}, 44\penalty0 (3), 2022.

\bibitem[Roessle et~al.(2022)Roessle, Barron, Mildenhall, Srinivasan, and Nie{\ss}ner]{roessle2022dense}
Barbara Roessle, Jonathan~T Barron, Ben Mildenhall, Pratul~P Srinivasan, and Matthias Nie{\ss}ner.
\newblock Dense depth priors for neural radiance fields from sparse input views.
\newblock In \emph{Proceedings of the IEEE/CVF Conference on Computer Vision and Pattern Recognition}, pages 12892--12901, 2022.

\bibitem[Sargent et~al.(2023)Sargent, Li, Shah, Herrmann, Yu, Zhang, Chan, Lagun, Fei-Fei, Sun, et~al.]{sargent2023zeronvs}
Kyle Sargent, Zizhang Li, Tanmay Shah, Charles Herrmann, Hong-Xing Yu, Yunzhi Zhang, Eric~Ryan Chan, Dmitry Lagun, Li Fei-Fei, Deqing Sun, et~al.
\newblock Zeronvs: Zero-shot 360-degree view synthesis from a single real image.
\newblock \emph{arXiv preprint arXiv:2310.17994}, 2023.

\bibitem[Sch\"{o}nberger and Frahm(2016)]{schoenberger2016sfm}
Johannes~Lutz Sch\"{o}nberger and Jan-Michael Frahm.
\newblock {Structure-from-Motion Revisited}.
\newblock In \emph{Conference on Computer Vision and Pattern Recognition (CVPR)}, 2016.

\bibitem[Sch\"{o}nberger et~al.(2016)Sch\"{o}nberger, Zheng, Pollefeys, and Frahm]{schoenberger2016mvs}
Johannes~Lutz Sch\"{o}nberger, Enliang Zheng, Marc Pollefeys, and Jan-Michael Frahm.
\newblock {Pixelwise View Selection for Unstructured Multi-View Stereo}.
\newblock In \emph{European Conference on Computer Vision (ECCV)}, 2016.

\bibitem[Seo et~al.(2023)Seo, Chang, and Kwak]{seo2023flipnerf}
Seunghyeon Seo, Yeonjin Chang, and Nojun Kwak.
\newblock Flipnerf: Flipped reflection rays for few-shot novel view synthesis.
\newblock In \emph{Proceedings of the IEEE/CVF International Conference on Computer Vision}, pages 22883--22893, 2023.

\bibitem[Song et~al.(2023)Song, Park, An, Cho, Kwak, Cho, and Kim]{song2023darf}
Jiuhn Song, Seonghoon Park, Honggyu An, Seokju Cho, Min-Seop Kwak, Sungjin Cho, and Seungryong Kim.
\newblock Därf: Boosting radiance fields from sparse inputs with monocular depth adaptation, 2023.

\bibitem[Sun et~al.(2022)Sun, Sun, and Chen]{sun2022direct}
Cheng Sun, Min Sun, and Hwann-Tzong Chen.
\newblock Direct voxel grid optimization: Super-fast convergence for radiance fields reconstruction.
\newblock In \emph{Proceedings of the IEEE/CVF Conference on Computer Vision and Pattern Recognition}, pages 5459--5469, 2022.

\bibitem[Sun et~al.(2023)Sun, Zhang, Chen, Li, Ji, Zhao, and Xing]{sun2023vgos}
Jiakai Sun, Zhanjie Zhang, Jiafu Chen, Guangyuan Li, Boyan Ji, Lei Zhao, and Wei Xing.
\newblock Vgos: Voxel grid optimization for view synthesis from sparse inputs.
\newblock In \emph{Proceedings of the Thirty-Second International Joint Conference on Artificial Intelligence, {IJCAI-23}}, pages 1414--1422. International Joint Conferences on Artificial Intelligence Organization, 2023.
\newblock Main Track.

\bibitem[Tang et~al.(2023)Tang, Ren, Zhou, Liu, and Zeng]{tang2023dreamgaussian}
Jiaxiang Tang, Jiawei Ren, Hang Zhou, Ziwei Liu, and Gang Zeng.
\newblock Dreamgaussian: Generative gaussian splatting for efficient 3d content creation.
\newblock \emph{arXiv preprint arXiv:2309.16653}, 2023.

\bibitem[Uy et~al.(2023)Uy, Martin-Brualla, Guibas, and Li]{uy2023scade}
Mikaela~Angelina Uy, Ricardo Martin-Brualla, Leonidas Guibas, and Ke Li.
\newblock Scade: Nerfs from space carving with ambiguity-aware depth estimates.
\newblock In \emph{Proceedings of the IEEE/CVF Conference on Computer Vision and Pattern Recognition}, pages 16518--16527, 2023.

\bibitem[Wang et~al.(2022)Wang, Wang, Zhang, Zhang, Bai, Ning, Zhou, and Hancock]{wang2022uncertainty}
Chen Wang, Xiang Wang, Jiawei Zhang, Liang Zhang, Xiao Bai, Xin Ning, Jun Zhou, and Edwin Hancock.
\newblock Uncertainty estimation for stereo matching based on evidential deep learning.
\newblock \emph{Pattern Recognition}, 124:\penalty0 108498, 2022.

\bibitem[Wang et~al.(2023)Wang, Chen, Loy, and Liu]{wang2023sparsenerf}
Guangcong Wang, Zhaoxi Chen, Chen~Change Loy, and Ziwei Liu.
\newblock Sparsenerf: Distilling depth ranking for few-shot novel view synthesis.
\newblock In \emph{Proceedings of the IEEE/CVF International Conference on Computer Vision (ICCV)}, pages 9065--9076, 2023.

\bibitem[Wang et~al.(2021)Wang, Wang, Liu, Zhou, Zhang, Zheng, and Bai]{wang2021multi}
Xiang Wang, Chen Wang, Bing Liu, Xiaoqing Zhou, Liang Zhang, Jin Zheng, and Xiao Bai.
\newblock Multi-view stereo in the deep learning era: A comprehensive review.
\newblock \emph{Displays}, 70:\penalty0 102102, 2021.

\bibitem[Wang et~al.(2024{\natexlab{a}})Wang, Luo, Wang, Zheng, and Bai]{wang2024robust}
Xiang Wang, Haonan Luo, Zihang Wang, Jin Zheng, and Xiao Bai.
\newblock Robust training for multi-view stereo networks with noisy labels.
\newblock \emph{Displays}, 81:\penalty0 102604, 2024{\natexlab{a}}.

\bibitem[Wang et~al.(2004)Wang, Bovik, Sheikh, and Simoncelli]{wang2004ssim}
Zhou Wang, Alan~C Bovik, Hamid~R Sheikh, and Eero~P Simoncelli.
\newblock Image quality assessment: from error visibility to structural similarity.
\newblock \emph{IEEE transactions on image processing}, 13\penalty0 (4):\penalty0 600--612, 2004.

\bibitem[Wang et~al.(2024{\natexlab{b}})Wang, Luo, Wang, Zheng, Ning, and Bai]{wang2024contrastive}
Zihang Wang, Haonan Luo, Xiang Wang, Jin Zheng, Xin Ning, and Xiao Bai.
\newblock A contrastive learning based unsupervised multi-view stereo with multi-stage self-training strategy.
\newblock \emph{Displays}, page 102672, 2024{\natexlab{b}}.

\bibitem[Wu et~al.(2023)Wu, Yi, Fang, Xie, Zhang, Wei, Liu, Tian, and Wang]{wu20234d}
Guanjun Wu, Taoran Yi, Jiemin Fang, Lingxi Xie, Xiaopeng Zhang, Wei Wei, Wenyu Liu, Qi Tian, and Xinggang Wang.
\newblock 4d gaussian splatting for real-time dynamic scene rendering.
\newblock \emph{arXiv preprint arXiv:2310.08528}, 2023.

\bibitem[Wu et~al.(2022)Wu, Li, Peng, Lu, Cao, and Zhong]{wu2022dof}
Zijin Wu, Xingyi Li, Juewen Peng, Hao Lu, Zhiguo Cao, and Weicai Zhong.
\newblock Dof-nerf: Depth-of-field meets neural radiance fields.
\newblock In \emph{Proceedings of the 30th ACM International Conference on Multimedia}, pages 1718--1729, 2022.

\bibitem[Wynn and Turmukhambetov(2023)]{wynn2023diffusionerf}
Jamie Wynn and Daniyar Turmukhambetov.
\newblock Diffusionerf: Regularizing neural radiance fields with denoising diffusion models.
\newblock In \emph{Proceedings of the IEEE/CVF Conference on Computer Vision and Pattern Recognition}, pages 4180--4189, 2023.

\bibitem[Xiong et~al.(2023)Xiong, Muttukuru, Upadhyay, Chari, and Kadambi]{xiong2023sparsegs}
Haolin Xiong, Sairisheek Muttukuru, Rishi Upadhyay, Pradyumna Chari, and Achuta Kadambi.
\newblock Sparsegs: Real-time 360° sparse view synthesis using gaussian splatting.
\newblock \emph{arXiv preprint arXiv:2312.00206}, 2023.

\bibitem[Xu et~al.(2023)Xu, Jiang, Wang, Fan, Wang, and Wang]{xu2023neurallift}
Dejia Xu, Yifan Jiang, Peihao Wang, Zhiwen Fan, Yi Wang, and Zhangyang Wang.
\newblock Neurallift-360: Lifting an in-the-wild 2d photo to a 3d object with 360deg views.
\newblock In \emph{Proceedings of the IEEE/CVF Conference on Computer Vision and Pattern Recognition}, pages 4479--4489, 2023.

\bibitem[Xu et~al.(2022)Xu, Xu, Philip, Bi, Shu, Sunkavalli, and Neumann]{xu2022point}
Qiangeng Xu, Zexiang Xu, Julien Philip, Sai Bi, Zhixin Shu, Kalyan Sunkavalli, and Ulrich Neumann.
\newblock Point-nerf: Point-based neural radiance fields.
\newblock In \emph{Proceedings of the IEEE/CVF Conference on Computer Vision and Pattern Recognition}, pages 5438--5448, 2022.

\bibitem[Yang et~al.(2023)Yang, Pavone, and Wang]{yang2023freenerf}
Jiawei Yang, Marco Pavone, and Yue Wang.
\newblock Freenerf: Improving few-shot neural rendering with free frequency regularization.
\newblock In \emph{Proceedings of the IEEE/CVF Conference on Computer Vision and Pattern Recognition}, pages 8254--8263, 2023.

\bibitem[Yu et~al.(2021{\natexlab{a}})Yu, Li, Tancik, Li, Ng, and Kanazawa]{yu2021plenoctrees}
Alex Yu, Ruilong Li, Matthew Tancik, Hao Li, Ren Ng, and Angjoo Kanazawa.
\newblock Plenoctrees for real-time rendering of neural radiance fields.
\newblock In \emph{Proceedings of the IEEE/CVF International Conference on Computer Vision}, pages 5752--5761, 2021{\natexlab{a}}.

\bibitem[Yu et~al.(2021{\natexlab{b}})Yu, Ye, Tancik, and Kanazawa]{yu2021pixelnerf}
Alex Yu, Vickie Ye, Matthew Tancik, and Angjoo Kanazawa.
\newblock pixelnerf: Neural radiance fields from one or few images.
\newblock In \emph{Proceedings of the IEEE/CVF Conference on Computer Vision and Pattern Recognition}, pages 4578--4587, 2021{\natexlab{b}}.

\bibitem[Yu et~al.(2022)Yu, Peng, Niemeyer, Sattler, and Geiger]{yu2022monosdf}
Zehao Yu, Songyou Peng, Michael Niemeyer, Torsten Sattler, and Andreas Geiger.
\newblock Monosdf: Exploring monocular geometric cues for neural implicit surface reconstruction.
\newblock \emph{Advances in neural information processing systems}, 35:\penalty0 25018--25032, 2022.

\bibitem[Zhang et~al.(2022{\natexlab{a}})Zhang, Yin, Wang, Yu, Fu, and Shen]{zhang2022hierarchical}
Chi Zhang, Wei Yin, Billzb Wang, Gang Yu, Bin Fu, and Chunhua Shen.
\newblock Hierarchical normalization for robust monocular depth estimation.
\newblock \emph{Advances in Neural Information Processing Systems}, 35:\penalty0 14128--14139, 2022{\natexlab{a}}.

\bibitem[Zhang et~al.(2022{\natexlab{b}})Zhang, Wang, Bai, Wang, Huang, Chen, Gu, Zhou, Harada, and Hancock]{zhang2022revisiting}
Jiawei Zhang, Xiang Wang, Xiao Bai, Chen Wang, Lei Huang, Yimin Chen, Lin Gu, Jun Zhou, Tatsuya Harada, and Edwin~R Hancock.
\newblock Revisiting domain generalized stereo matching networks from a feature consistency perspective.
\newblock In \emph{Proceedings of the IEEE/CVF Conference on Computer Vision and Pattern Recognition}, pages 13001--13011, 2022{\natexlab{b}}.

\bibitem[Zhang et~al.(2018)Zhang, Isola, Efros, Shechtman, and Wang]{zhang2018lpips}
Richard Zhang, Phillip Isola, Alexei~A Efros, Eli Shechtman, and Oliver Wang.
\newblock The unreasonable effectiveness of deep features as a perceptual metric.
\newblock In \emph{Proceedings of the IEEE conference on computer vision and pattern recognition}, pages 586--595, 2018.

\bibitem[Zhou et~al.(2016)Zhou, Tulsiani, Sun, Malik, and Efros]{zhou2016view}
Tinghui Zhou, Shubham Tulsiani, Weilun Sun, Jitendra Malik, and Alexei~A Efros.
\newblock View synthesis by appearance flow.
\newblock In \emph{Computer Vision--ECCV 2016: 14th European Conference, Amsterdam, The Netherlands, October 11--14, 2016, Proceedings, Part IV 14}, pages 286--301. Springer, 2016.

\bibitem[Zhou et~al.(2023)Zhou, Wang, Zheng, and Bai]{zhou2023Adaptive}
Xiaoqing Zhou, Xiang Wang, Jin Zheng, and Xiao Bai.
\newblock Adaptive spatial sparsification for efficient multi-view stereo matching.
\newblock \emph{Acta Electronica Sinica}, 51\penalty0 (11):\penalty0 3079--3091, 2023.

\bibitem[Zhou and Tulsiani(2023)]{zhou2023sparsefusion}
Zhizhuo Zhou and Shubham Tulsiani.
\newblock Sparsefusion: Distilling view-conditioned diffusion for 3d reconstruction.
\newblock In \emph{CVPR}, 2023.

\bibitem[Zhu et~al.(2023)Zhu, Fan, Jiang, and Wang]{zhu2023fsgs}
Zehao Zhu, Zhiwen Fan, Yifan Jiang, and Zhangyang Wang.
\newblock Fsgs: Real-time few-shot view synthesis using gaussian splatting.
\newblock \emph{arXiv preprint arXiv:2312.00451}, 2023.

\end{thebibliography}
}


\end{document}